\documentclass[smallextended]{svjour3}

\usepackage{epsfig,subfigure,epstopdf}
\usepackage{amsfonts}
\usepackage{amssymb}
\usepackage{graphicx,psfrag,epsf}
\usepackage{amsmath}
\usepackage{enumerate}
\usepackage{natbib}
\usepackage{url} 
\usepackage{hyperref}
\usepackage{soul}
\newcommand{\blind}{0}

\newcommand{\bfmu}{\boldsymbol {\mu}}

\usepackage[usenames,dvipsnames]{color}

\begin{document}

\def\spacingset#1{\renewcommand{\baselinestretch}%
{#1}\small\normalsize} \spacingset{1}


\if0\blind
{
  \title{
  {Finding Optimal Points for}  Expensive Functions Using Adaptive RBF-Based Surrogate Model Via Uncertainty Quantification}

 \titlerunning{Adaptive RBF-Based Surrogate Model}

 \author{Ray-Bing Chen$^{*}$ \and
         Yuan Wang$^{*}$\thanks{$^{*}$ Joint first authors.} \and
  C. F. Jeff Wu$^{\dag}$\thanks{$^{\dag}$ Corresponding author}
   }

\institute{R.-B. Chen \at
Department of Statistics,
National Cheng Kung University \and
Y. Wang \at
Wells Fargo \and
C. F. J. Wu \at
H. Milton Stewart School of Industrial \& Systems Engineering, \\
   Georgia Institute of Technology}

  \maketitle
} \fi

\if1\blind
{
  \bigskip
  \bigskip
  \bigskip
  \begin{center}
    {\LARGE\bf Finding Optimal Points for Expensive Functions Using Adaptive Radial Basis Functions  Based Surrogate Model Via Uncertainty Quantification}
\end{center}
  \medskip
} \fi

\begin{abstract}
Global optimization of expensive functions has important applications in physical and computer experiments.
It is a challenging problem to develop efficient optimization scheme, because each function evaluation can be costly and the derivative information of the function is often not available. 
We propose a novel global optimization framework using adaptive Radial Basis Functions (RBF) based surrogate model via uncertainty quantification. The framework consists of two iteration steps. It first employs an RBF-based Bayesian surrogate model to approximate
the true function, where the parameters of the RBFs can be adaptively estimated and updated each time a new point is explored.
Then it utilizes a model-guided selection criterion to identify a new point from a candidate set for function evaluation.
The selection criterion adopted here is a sample version of the expected improvement (EI) criterion.
We conduct simulation studies with standard test functions, which show that the proposed
method 
{has some advantages, especially when the true surface is not {very} smooth. In addition, we also {propose} modified approaches to improve the search performance for identifying global optimal points and to deal with the higher dimension scenarios.}
\end{abstract}




\section{Introduction}

In this paper, we consider the problem of global optimization of
expensive functions, i.e., functions which require large
computational costs to evaluate. For physical and computational
experiments, these functions represent the relationship between
input and output variables, and may require days or even weeks to
evaluate at a single input setting. One example is the
high-pressure mixing and combustion processes in liquid rocket
engines, which requires numerically solving a large, coupled system
of partial differential equations; see \cite{oefelein1998}.
Even when computation is parallelized over thousands of processing
cores, a comprehensive simulation of a single injector may take months
to complete. An important problem about expensive functions is how
to optimize the output/response by choosing appropriate settings of
the input variables. This problem can be challenging for two
reasons. First, it is not feasible to conduct extensive runs of
function evaluations to find the optimal input settings, since each
function evaluation is expensive. It is thus desirable to identify
the optimal input settings with as few runs as possible. The second
challenge comes from the complicated nature of the functional
relationship. They are usually regarded as ``black-boxes'', because
there is no explicit relationship between the input and output.
Although various local optimization methods are available when the
derivatives of the functions are known or can be easily obtained,
see \cite{boyd2004convex}, such methods are not applicable in the
present scenario.

In the literature, a widely used practice for global optimization of
expensive functions is to sequentially select input settings for
function evaluations based on some criterions. The approach consists
of two steps. First, it constructs a surrogate model to approximate
the true function based on all the observed function outputs. The
advantage of employing  surrogate model is that it  can provide
predictions at any input settings with much cheaper computation.
Second, it identifies a new input setting for function evaluation
according to some surrogate model based selection criteria. With
this approach, it is feasible to approximate the global optimizer of
the true function via the surrogate model optimization. {The commonly used surrogate models are the kriging {models} \citep{jones1998efficient, jones2001taxonomy} and { models based} on radial basis functions \citep{gutmann2001radial,regis2007stochastic}. \cite{chen2010building, chen2016sequential} proposed to construct the surrogate models via overcomlpete pre-specified basis functions.
In addition, another type of optimization approach is statistical global optimization {which chooses the next point} based on a probability improvement criterion, like P-algorithm \citep{Torn:1989:GO:75021}. \cite{gutmann2001radial} and \cite{zilinskas2010}
have showed the equivalence of the P-algorithm and the surrogate approach proposed in \cite{gutmann2001radial} under certain conditions. For more details along these lines, see a review in \cite{zilinskas2016}.}

The primary objective of this paper is to propose a novel global
optimization framework for optimizing expensive functions. Our
approach is motivated by \cite{regis2007stochastic}, in which they
utilize Radial Basis Functions (RBF) to build a deterministic
surrogate model and guide the selection of the next explored point
based on the predicted response and some distance criteria. The
rationale of using RBFs is that they can capture the nonlinear trend
of functions. However, the RBFs they used are pre-determined and
lack the flexibility of modeling. Also, it is less efficient to
perform function evaluation from their surrogate model, because they
use RBFs in a deterministic way without providing prediction
uncertainties. Although a distance criterion is used to avoid
{getting trapped} at local optima, it does not incorporate {the
information in response values for the surrogate models. To make
better use of all information in the data,} we propose to construct
surrogate model with RBFs that are chosen adaptively based on the
updated outputs, and to select new points based on surrogate models
with quantified uncertainties.

There are other approaches for global optimization of expensive
functions in the literature. \cite{jones1998efficient} propose a
global optimization scheme by constructing a surrogate model with
the kriging method. Our approach is different in that they make
strong assumptions on the correlation structure between explored
points while ours does not. A detailed review related to the kriging
model in global optimization can be found in
\cite{jones2001taxonomy}. \cite{chen2016sequential} propose a global
optimization scheme that builds a mean prediction model with linear
basis functions selected from a dictionary of functions, and then
imposes a Bayesian structure over the mean model to quantify the
uncertainty of the prediction. Our approach is also different from
\cite{chen2016sequential}. Instead of using a predetermined discrete
function dictionary with a large number of linear functions, we use
a moderate number of RBFs that can be adaptively updated based on
observed data.

The paper is organized as follows. In Section \ref{sec:form_RBF}, we
give a mathematical formulation of the global optimization problem,
and provide a review of the RBFs. In Section \ref{sec:framework}, we
present the proposed global optimization framework that utilizes
adaptive RBF-based Bayesian surrogate model.
In Section \ref{sec:simulation}, we present simulation studies to
validate and compare our proposed method with the methods  by
\cite{regis2007stochastic} and \cite{jones1998efficient}. Further simulation results for 3- and
4-dimensional functions are given in Section \ref{sec:higher_dim}.
A modification of the proposed method to avoid getting trapped in local optima is presented in Section \ref{sec:M-aBRF}.
{In addition, we study the effect of the grid size which is used as the candidate points in the proposed method, and also come out a grid-free approach.
}
Concluding remarks and future research directions are given in
Section \ref{sec:conclusion}.


\section{Problem Formulation and Review of RBFs} \label{sec:form_RBF}

Suppose $f({\bf x})$ is an expensive function of interest, where ${\bf x}  = (x^1, ..., x^p)^T  \in V,$ and $V$ is a {$p$-dimensional} convex domain {in $R^p$}.
The objective is to  identify an optimal input setting ${\bf x}_{opt}$ that maximizes $f({\bf x}),$
\begin{eqnarray}
{\bf x}_{opt} = \arg\max_{{\bf x} \in V} f({\bf x}).
\end{eqnarray}
Because it is not practical to evaluate $f({\bf x})$ over $V$ to
search the  global maximizer due to the huge computational cost, a
well-established practice is to sequentially select a few input
settings for function evaluation using a two-step strategy. Suppose
a set of $N$ function evaluations $\{({\bf x}_i, f({\bf
x}_i))\}_{i=1}^{N}$ are taken. In step 1,  a surrogate model is
constructed and the resulting model approximation is denoted by
$f_N({\bf x}).$ {Unlike the true function $f({\bf x}),$ the surrogate model is much cheaper to build and evaluate. Therefore it is feasible to predict function values over all ${\bf x} \in V.$}
In step 2, the next input setting ${\bf x}_{N+1}$ is selected
for function evaluation via certain criterion based on  the
surrogate model  from step 1. Steps 1 and 2 iterate until the total
computational budget is met.
{The best point among {all the chosen input settings,}
$\hat{\mathbf{x}}_{opt} = \arg\max_{\mathbf{x}_i} f(\mathbf{x}_i)$, can be
treated as an approximation to the true optimal point
$\mathbf{x}_{opt}$.}
By adopting
this two-step strategy, we will present in Section \ref{sec:framework} our proposed framework in detail.
Note that the surrogate construction may not necessarily be an interpolator of the observed points, i.e., $f_N({\bf x}_i) \neq f({\bf x}_i).$ Because our goal is optimization, the surrogate is used to predict the location of the optimal points, rather than to approximate the response with high accuracy \citep{chen2010building}.
Thus we want to capture the trend of the true response surface quickly and to serve this purpose, our surrogate model does not have to meet the interpolation requirement.

In the remaining part of this section, we give a brief review of
the RBFs, which will be used  in the proposed framework for the
surrogate model construction. In the literature, the RBF is
popularly deployed in applied mathematics and neural networks. See
\cite{buhmann2003radial} and \cite{bishop2006pattern}.
Several commonly used functions are: (1) Gaussian functions: $r({\bf
x}; \bfmu, s) = \exp\{-s^2||{\bf x}-\bfmu||^2\};$ (2) generalized
multi-quadratic functions: $r({\bf x}; \bfmu, s) = (||{\bf
x}-\bfmu||^2 + s^2)^{\beta}$ with $s>0, 0<\beta<1;$ (3) generalized
inverse multi-quadratic functions: $r({\bf x}; \bfmu, s) = (||{\bf
x}-\bfmu||^2 + s^2)^{-\beta}$ with $s>0, \beta>0;$ (4) thin plate
spline functions: $r({\bf x}; \bfmu) = ||{\bf x}-\bfmu||^2 \ln
(||{\bf x}-\bfmu||),$
{{where} $\mu$ is the center of the function, and $s$ is a pre-specified constant {which varies with the chosen function.}} 
In our work, we will focus on the Gaussian
RBFs. The Gaussian RBFs have two types of parameters: the center
parameter $\bfmu \in V$ that determines the location of the RBFs,
and the scale parameter $s$ that measures the degree of fluctuation
of the function. One advantage of using the Gaussian RBFs over other
basis functions is that it can capture different trends of response
by choosing different centers and scales. For example, a larger $s$
indicates a more concentrated change in the surface, and vice versa.

\section{General Global Optimization Framework} \label{sec:framework}

In this section, we propose a global optimization framework that
utilizes adaptive RBF-based surrogate model via uncertainty
quantification. In Section \ref{sec:surrogate-model}, we propose a
novel hierarchical normal mixture Bayesian surrogate model with RBFs
to approximate the true function, where the model coefficients are
sparsely represented to avoid over-fitting, and the parameters of
the RBFs are adaptively updated each time a new point is explored.
This allows us to predict the function value at any given candidate
point. In Section \ref{sec:select_point}, we propose a model-guided
selection criterion and based on the posterior samples, a sample version of the expected improvement criterion is adopted. A new point can then
be selected to identify a more promising area of global
maximizer. 
A summary of the algorithm and some discussions will be presented in
Section \ref{sec:algo&rem}.

\subsection{Normal Mixture Surrogate Model with RBFs} \label{sec:surrogate-model}

Suppose we observe $N$ explored points $\mathcal{P}_{exp} =\{{\bf
x}_1, \dots,{\bf x}_N \},$ and its function values $y  = (y_1, ...,
y_N)^T = (f({\bf x}_1), ..., f({\bf x}_N))^T$.
Without loss of generality, we assume $E(y_i)=0,$ {because}
otherwise we can approximate $(y_i - \bar{y})$'s instead of $y_i$'s, {where $\bar{y}$ is the sample mean of $y_1, \dots, y_N$, i.e., $\bar{y} = \sum_{i=1}^{N} y_i / N$.}
We propose to construct a surrogate model by a summation of $N$
Gaussian RBFs $r({\bf x}; \bfmu_i, s_i) =  \exp\{-s_i^2||{\bf
x}-\bfmu_i||^2\}$ and an error term $\epsilon({\bf x})$:
\begin{eqnarray} \label{eq:f_hat}
f({\bf x}) = f_N({\bf x}) +  \epsilon({\bf x}) = \sum_{i=1}^{N} \beta_{i} r({\bf x}; \bfmu_i, s_i) + \epsilon({\bf x}).
\end{eqnarray}
Here, an error term is used to model the discrepancy between the
model approximation $f_N({\bf x})$ constructed by the RBFs and the
true function $f(x).$ We assume that  $\epsilon({\bf x})$
{are independent normal distributions with mean 0 and vairance $\sigma^2$.}
Note that if the center parameters $\bfmu_i$'s and
the scale parameters $s_i$'s are known and fixed, then the surrogate
model in (\ref{eq:f_hat}) is exactly the same as linear regression.

\subsubsection{Prior Distributions}

Because both $\bfmu_i$'s and $s_i$'s are  unknown, the proposed modeling approach can handle highly nonlinear functions.
A uniform prior over a rectangular region is used for  $\bfmu= (\bfmu_1, ..., \bfmu_N),$
\begin{eqnarray}
\bfmu_i \sim \text{Uniform}(\Omega), i=1,...,N,
\end{eqnarray}
where $\Omega = \prod_{j=1}^p[\min(x_{1:N}^j), \max(x_{1:N}^j))],$ {i.e., the smallest hyper-rectangle to cover the current explored points,}
and it is adaptively changed with the addition of  new explored
points, see  \cite{andrieu2001robust}. A gamma prior is used for the
scale parameters $s = (s_1, ..., s_N)^T,$
\begin{eqnarray} \label{eq:gamma_prior}
s_i \sim \text{Gamma}(a_s, b_s),
\end{eqnarray}
where $a_s$ and $b_s$ are common to all $i$'s.

We also impose a hierarchical structure on the coefficients
$\beta_i$'s. Define a latent variable $\gamma = (\gamma_1,...,
\gamma_{N})^T$ to indicate whether a certain basis function is
{important} or not: $\gamma_i = 1$ indicates that the $i$th basis is
important, while $\gamma_i = 0$ indicates the opposite.
Specifically,  we set $\beta_i | (\gamma_i=0) \sim \text{\it N}(0,
\tau_{i})$ with small $\tau_i,$ and $\beta_i | (\gamma_i=1) \sim
\text{\it N}(0, C \tau_{i})$ with relatively large $C$, {where $C$
can be interpreted as a variance ratio.} This hierarchical setting
is first employed in the Stochastic Search Variable Selection (SSVS)
scheme  by \cite{george1993variable} {and is also used for
uncertainty quantification studies in \cite{chen2016sequential}.}
Indeed, it is one type of the ``g-prior'' (see
\cite{zellner1986assessing}) for avoiding over-fitting. Now the
mixture normal prior of the model coefficient $\beta = (\beta_1,
..., \beta_N)^T$ can be written as follows:
\begin{eqnarray} \label{eq:beta}
\beta| \gamma \sim N(0, \Sigma_\tau^2), \text{ where } \Sigma_\tau =\text{diag}(a_1 \tau_1 ,..., a_{N} \tau_{N}),
\end{eqnarray}
with $a_i = 1$ if $\gamma_i = 0$ and $=C$ if $\gamma_i = 1,$
and a binomial prior  for the latent variable $\gamma_i,$
\begin{eqnarray}
P(\gamma_i = 0) = p_i, P(\gamma_i = 1) = 1-p_i, i=1,...,N.
\end{eqnarray}
{Note that the choice of $C$ plays an important role in the
posterior sampling and control the model complexity.} We also impose
an inverse-gamma prior  for the residual variance $\sigma^2,$
\begin{eqnarray} \label{eq:sigma_prior}
\sigma^2 \sim IG(\frac{\nu_0}{2}, \frac{\zeta_0}{2}).
\end{eqnarray}

By combining (\ref{eq:f_hat})-(\ref{eq:sigma_prior}) {with
independent prior assumptions}, we obtain the full posterior
distribution of $\{\beta, \bfmu, \gamma, \sigma^2, s\}$
\begin{eqnarray}  \label{eq: full}
& &p(\beta, \bfmu, \gamma, \sigma^2, s | {\bf x}, y) \propto p(y| \beta, \bfmu, \gamma, \sigma^2,s, {\bf x})  \cdot p(\beta|\gamma, \bfmu) \cdot p(\gamma) \cdot p(s)  \cdot p(\bfmu) \cdot p(\sigma^2)    \cr
&=&
\Big [ (2\pi\sigma^2)^{-N/2} \exp \{-\frac{1}{2\sigma^2}(y-D(\bfmu,s) \cdot \beta)^T (y-D(\bfmu,s) \cdot \beta)\} \Big ] \Big[ \prod_{i=1}^{N+p} p_i^{\gamma_i}(1-p_i)^{(1-\gamma_i)}\Big]\cr
& &\Big[ \det(2\pi \Sigma_\tau^2)^{-1/2}  \exp\{-\frac{1}{2} \beta^T \Sigma_\tau^{-2} \beta\} \Big]
 \prod_{i=1}^{N}\Big[\frac{b_s^{a_s}}{\gamma(a_s)}s_i^{a_s - 1}\exp(-b_s s_i) \Big]
\Big[ \frac{1_{\Omega}(\bfmu_{1:N})}{V(\Omega)} \Big] \cr
& &\Big[ (\sigma^2)^{-(\nu_0/2+1)} \exp\{-\frac{\zeta_0}{2\sigma^2}\}\Big],
\end{eqnarray}
where the coefficient matrix $D(\bfmu, s) $ is defined as
\begin{eqnarray*}
D(\bfmu, s) =
 \begin{pmatrix}
   r({\bf x}_1; \bfmu_1, s_1) & \cdots & r({\bf x}_1;  \bfmu_N, s_N) \\
  \vdots & \ddots & \vdots \\
   r({\bf x}_N; \bfmu_1, s_1) & \cdots & r({\bf x}_N; \bfmu_N, s_N)
 \end{pmatrix},
\end{eqnarray*}
and the indicator function $1_{\Omega}(x)=1$ if $x \in \Omega,$  $=0$ if $x \notin \Omega.$

\subsubsection{Posterior Sampling}
The posterior distribution defined in (\ref{eq: full}) is
computationally intractable. Markov Chain Monte Carlo (MCMC) method
is utilized to solve this problem, see \cite{andrieu2001robust} and
\cite{koutsourelakis2009accurate}.
{That is{,} we use the MCMC method to generate the posterior samples from $p(\beta, \bfmu, \gamma, \sigma^2, s | {\bf x}, y)$. Thus we sequentially sample $\beta$ $\gamma$, $\sigma^2$, $\bfmu$ and $s$ by fixing the other components and the data ${\bf x}, y$. {Under} certain conditions, we can guarantee that these samples can be treated as the posterior samples of $\beta, \gamma, \sigma^2, \bfmu, s$.}
{Here the MCMC method iterates the following two steps:
\begin{itemize}
\item Sample $\beta, \gamma, \sigma^2$ by fixing $\bfmu$, $s$, $\mathbf{x}$ and $y$.
\item Sample $\bfmu$ and $s$ by fixing $\beta$, $\gamma$, $\sigma^2$, $\mathbf{x}$ and $y$.
\end{itemize}}
\noindent Specifically, we use the Gibbs
sampler to generate the posterior samples for the parameters
$\beta, \gamma, \sigma^2,$ and the Metropolis-Hasting algorithm to
{obtain the posterior samples} for the  parameters $\bfmu$ and
$s,$ because there is no explicit formula for the posterior
distributions of $\bfmu$ and $s$.

Start with the posterior
distributions for $\beta,  \gamma, \sigma^2.$ Denote $M = (D(\bfmu,
s)^T D(\bfmu, s)/\sigma^2+ \Sigma_\tau^{-2})^{-1},$ and $h = MD(\bfmu, s)^T
y/\sigma^2$. 
Then,
the samples of $\beta$ can be generated by
\begin{eqnarray} \label{eq:post_beta}
\beta | \bfmu, \sigma^2, \gamma, s, {\bf x}, y \sim \text{N}(h, M).
\end{eqnarray}
The samples of $\sigma^2$ can be generated by
\begin{eqnarray}
\sigma^2 | \beta, \bfmu, \gamma, s, {\bf x}, y \sim \text{IG}(\frac{\nu_0 +N}{2},\frac{\zeta_0+|y-D(\bfmu,s) \beta|^2}{2}).
\end{eqnarray}
{For the  samples of $\gamma$, it would be simple to sample $\gamma_i$ sequentially conditional on the other components, and} $\gamma_i$ can be generated by
\begin{eqnarray}
P(\gamma_i = 1|\beta, \bfmu, \sigma, \gamma_{(-i)}, {\bf x}, y) = p_1/(p_1+p_0),
\end{eqnarray}
where
\[
p_1 = f(\beta| \gamma_i = 1, \gamma_{-i}, \bfmu)f(\gamma_i = 1, \gamma_{-i}) \propto \det(\Sigma^*)^{-1/2} \exp \{ -\frac{1}{2}\beta^T(\Sigma^*)^{-1}\beta \} (1-p_i)
\]
with $\Sigma^* = D_r^{i+},$ and $D_r^{i+}$ is $\Sigma_\tau$ with $\gamma_i = 1$;
\[
p_0 = f(\beta| \gamma_i = 0, \gamma_{-i}, \bfmu)f(\gamma_i = 0, \gamma_{-i}) \propto \det(\Sigma^*)^{-1/2} \exp \{ -\frac{1}{2}\beta^T(\Sigma^*)^{-1}\beta \} p_i
\]
with $\Sigma^* = D_r^{i-},$ and $D_r^{i-}$  is $\Sigma_\tau$ with $\gamma_i = 0.$
{Here} the notation $\gamma_{-i} = (\gamma_1, ..., \gamma_{i-1}, \gamma_{i+1}, ..., \gamma_N)^T$ represents the vector of all $\gamma$'s except $\gamma_i.$

Now we turn to 
the parameters $\bfmu$ and $s.$
{First, consider the {sampling} procedure for $\bfmu$. Instead of directly sampling the vector $\bfmu$, we suggest sampling $\bfmu_i$ sequentially from}
\begin{eqnarray} \nonumber  & & \label{eq:post_mu}
p(\bfmu_i | \bfmu_{-i}, \beta, s, \sigma, {\bf x}, y) \\ 
&\propto& \exp \{-\frac{1}{2\sigma^2}(y-D(\bfmu,s)  \beta)^T (y-D(\bfmu,s) \beta)\} 1_{\Omega}(\bfmu_{1:N}),
\end{eqnarray}
where $\bfmu_{-i} = (\bfmu_1, ..., \bfmu_{i-1}, \bfmu_{i+1}, ..., \bfmu_N)$ denotes the vector of all $\bfmu$'s except $\bfmu_i.$
We use the Metropolis-Hasting algorithm to generate posterior samples for $\bfmu_i.$
Specifically, at a new step $(k+1)$, we set the proposed density to be a mixture of two densities, and  a temporary sample $\bfmu_i^{*}$ can be obtained from the  whole domain $\Omega$ with uniform probability,
or it can be  a perturbation  of the current  iteration $\bfmu_i^{(k)}$ within its local neighborhood,  i.e.,
\begin{eqnarray} \label{eq: proposed}
& & q_1(\bfmu_i^{*}) = \text{Uniform}(\Omega), \text{ with probability } \omega,  \cr
\text{ and } & & q_2(\bfmu_i^{*}) = N(\bfmu_i^{(k)}, \sigma_{\mu}^2) \text{ with probability } 1-\omega.
\end{eqnarray}
{Then} we accept this temporary sample $\bfmu_i^*$ with the acceptance rate
\[
A(\bfmu_i,\bfmu_i^*) = \min\{1, (\frac{\exp \{-1/(2\sigma^2)|y-D(\bfmu^*,s)  \beta|^2\} }{\exp \{-1/(2\sigma^2)|y-D(\bfmu,s) \beta|^2\} }) 1_{\Omega}(\bfmu_1, .., \bfmu_i^*,...,\bfmu_N)\}
\]
where $\bfmu^*=(\bfmu_1, ..., \bfmu_i^*, ..., \bfmu_N)^T.$


Similarly, we can use the Metropolis-Hasting algorithm to generate samples of $s_i.$
At step $(k+1),$ we choose a temporary $s_i^*$ as a perturbation of the current sample $s_i^{(k)}$ by the proposed density
\begin{eqnarray} \label{eq: proposed3}
 q_3(s_i^{*}) = N(s_i^{(k)}, \sigma_s^2).
\end{eqnarray}
And we accept such sample $s_i^*$ with the acceptance rate
\[
A(s_i, s_i^*) = \min\{1, (\frac{\exp \{-1/(2\sigma^2)|y-D(\bfmu,s^*)  \beta|^2\} }{\exp \{-1/(2\sigma^2)|y-D(\bfmu,s) \beta|^2\} }\cdot \frac{(s_i^*)^{a_s - 1}\exp(-b_s s_i^*)}{s_i^{a_s - 1}\exp(-b_s s_i)})\}
\]
where $s^*=(s_1, ..., s_i^*, ..., s_N).$

From (\ref{eq:post_beta})--(\ref{eq: proposed3}), we generate samples for $\gamma, \beta, \sigma, \bfmu, s$ iteratively based the updated estimate for the remaining parameters. 
Then, the Gibbs sequence
$$ \gamma^{(0)}, \beta^{(0)}, \sigma^{(0)}, \bfmu^{(0)}, s^{(0)},...,  \gamma^{(k)}, \beta^{(k)}, \sigma^{(k)},\bfmu^{(k)}, s^{(k)},... , \gamma^{(K)}, \beta^{(K)}, \sigma^{(K)},\bfmu^{(K)}, s^{(K)}$$
can be obtained, where $K$ is the total number of iterations.
After discarding the first say 40\% samples, {the remaining samples can be treated as the posterior samples of $\beta$, $\gamma$, $\sigma^2$, $\bfmu$ and $s$ from $p(\beta, \bfmu, \gamma, \sigma^2, s | {\bf x}, y)$.
Thus}
the posterior sample  $f_N^{(k)} (\tilde{{\bf x}})$ for model approximation at a candidate explored point $\tilde{{\bf x}}$ can be calculated by
\begin{eqnarray} \label{post_pred}
f_N^{(k)} (\tilde{{\bf x}})=  \sum_{i=1}^{N} \beta_{i}^{(k)} r(\tilde{{\bf x}}; \bfmu_i^{(k)}, s_i^{(k)}).
\end{eqnarray}
Then the function prediction $f_N(\tilde{{\bf x}})$ can then be calculated as the average of $f_N^{(k)}(\tilde{{\bf x}})$'s, and the prediction uncertainties can be calculated as the sample variance of $f_N^{(k)}(\tilde{{\bf x}})$'s.

Finally, we note that the mean value of the posterior density of $\beta$ in (\ref{eq:post_beta}) is $h = ((D(\bfmu, s)^T \cdot$ $D(\bfmu, s)/\sigma^2+ \Sigma_\tau^{-2})^{-1} )D(\bfmu, s)^T y/\sigma^2,$ which is a biased estimator of $\beta$ with a nugget value $\Sigma_\tau^{-2}.$ Hence, this estimate of $\beta$ can be regarded as a ridge-type regression estimate.
It is deployed to prevent the model coefficients from being too large. Its use can lead to a more stable  surrogate model.

\subsubsection{Tuning Parameters} \label{sec:turning-par}
A remaining issue in the Bayesian computation is the tuning of the
hyper-parameters, which is critical for the model performance. For
the hyper-parameters related to the RBF, we adopt the settings in
\cite{andrieu2001robust} and \cite{koutsourelakis2009accurate}.
Specifically, for the proposed density of the RBF centers $\bfmu_i$
in (\ref{eq: proposed}), we set  $\sigma_{\mu}^2 = 0.001.$ For the
prior of the RBF scales $s_i$ in (\ref{eq:gamma_prior}), we set $a_s
= 2, b_s = 0, $ and  for the proposed density of $s_i$ in (\ref{eq:
proposed3}), we set $\sigma^2_s = 0.5.$ For the hyper-parameters
related to model coefficients and residuals, we follow the settings
in \cite{chipman1997bayesian}. Specifically, for $\tau_i$, we
suggest to set $\tau_i = \Delta y / (3 \Delta {\bf x})$, where
$\Delta {x} = \max({\bf x}_{1:N}^{1:p}) - \min({\bf
x}_{1:N}^{1:p}),$ i.e., the largest change in ${\bf x}_{1:N},$ and
$\Delta y = \sqrt{Var(y)}/5$. For the prior of the  indicator
variable $\gamma_i$, we set $p_i = 0.5,$ i.e., the probability of
selecting a variable  is 50\%. For the hyper parameter $\nu_0$ and
$\gamma_0$ in (\ref{eq:sigma_prior}), we set $\nu_0 = 2,$ and
$\nu_0\gamma_0$ to be the 99\% quantile of the inverse gamma prior
that is close to $\sqrt{Var(y)}.$ {Consider the variance ratio $C$.
Usually we choose a large positive value for $C$, e.g., $C \geq 10$.
From our experience, we fix $C = 25$ in the first simulation example. However, it should be
problem-dependent and can be changed for different optimization
problems.}

\subsection{A Point Selection Criterion} \label{sec:select_point}

In this section, we discuss how to select new explored points based on the uncertainty of the response prediction for exploring uncertain regions. The ideal selection criterion should perfectly balance between exploration and exploitation properties to efficiently identify the optimal points within the given search budget. Here a sample version of the Expected Improvement criterion is adopted.


The EI criterion, initially proposed by \cite{Mockus1978},  is used to select points close to the global maxima based on a chosen surrogate model. Using this criterion, an explored point is selected to maximize the expected improvement over the best observed response
\begin{eqnarray}
E(I({\bf x})) = E(\max\{y- f_{\max}, 0\}),
\end{eqnarray}
where $f_{\max} = \max\{y_1, ..., y_N\}$ is the maximum of the observed model outputs.
It is pointed out in \cite{jones1998efficient} that under the Gaussian assumption of $y\sim N(\mu,  s_0^2),$ $E(I({\bf x}))$ has the following closed form expression:
\begin{eqnarray}
E(I({\bf x}))  = (\mu - f_{\max}) \Phi(\frac{\mu - f_{\max}}{s_0}) + s_0 \phi(\frac{\mu - f_{\max}}{s_0}).
\label{EI_n}
\end{eqnarray}
By examining the terms, we see that the expected improvement is large for those ${\bf x}$ having either (i) a predicted value at ${\bf x}$ that is much larger than the maximum of outputs obtained so far, i.e., $\mu \gg f_{\max},$ {or} (ii) having much uncertainty about the value of $y({\bf x}),$ i.e., when $s_0$ is large.

In our scenario, since the proposed surrogate model does not satisfy the Gaussian assumption,  there is no analytical form for $y,$ and thus it is not practical to calculate  $E(I({\bf x}))$ directly. Instead, we calculate the {\it Sampled Expected Improvement} (SEI) as suggested in {\cite{chipman2012} and \cite{chen2016sequential}}, i.e., to estimate $E(I({\bf x}))$ based on the posterior samples of $y,$
\begin{eqnarray}  \label{eq: EI}
\hat{E}(I({\bf x})) = \sum_{m=1}^{M} (\max\{y^{(m)}({\bf x}) -
f_{\max}, 0\})/M,
\end{eqnarray}
where $y^{(m)}({\bf x}) = f_N^{(m)}(x)$ is the $m$th posterior sample by (\ref{post_pred}), and $M$ is the total number of posterior samples.
Unlike in the Gaussian case, the SEI value in (\ref{eq: EI}) may not be expressed as a weighted sum of the improvement term and the prediction uncertainty term. From its definition, only the prediction posterior samples $y^{(m)}(x)$ that are larger than the current best value, $f_{max}$, are taken in the summation. Thus SEI first identifies the possible ``improvement’'' area, $\{x|y^{(m)}(x) > f_{max} \text{ for some } m\}$, and then sums over these terms.

A new explored point ${\bf x}_{N+1}$ at step $N+1$ is then selected to maximize the SEI criterion $\hat{E}(I({\bf x}))$,
\begin{eqnarray} \label{eq:arg_max_sn}
{\bf x}_{N+1} = \arg\max_{{\bf x} \in V \setminus P_{exp}} \hat{E}(I({\bf x})),
\end{eqnarray}
{where $P_{exp}$ is the current explored point set.} {Since the SEI criterion does not have a closed form, it may not be easy to identify the next explored point by solving (\ref{eq:arg_max_sn}). Thus to identify the next point, we may limit the search over a pre-specified grid, $\chi$, instead of {over} the whole region, $V$. }

\subsection{The Proposed Algorithm and Remarks} \label{sec:algo&rem}
In the first part of this section, we will present a summary of the
algorithm and the flexible usage of the proposed adaptive RBF-based
global optimization framework. For abbreviation, we will refer to
the proposed method as {BaRBF, where ``Ba'' stands for
``Bayesian adaptive''.} In the second part, we will compare our method with
the baseline method proposed in \cite{regis2007stochastic}.

Algorithm 1 summarizes the proposed global optimization method. {In
the beginning, the initial design is chosen as a space-filling
design. In this paper, the maximin Latin hypercube design
\citep{morris95} is used. Then the main body of Algorithm 1 is to
iterate between the two steps for the surrogate model construction
in Section \ref{sec:surrogate-model} and the point selection
criterion in Section \ref{sec:select_point}.}
\begin{table}[t]
\centering
\label{algo}
\begin{tabular}{ll}
\hline
\multicolumn{2}{l}{{\bf Algorithm 1} Global Optimization Algorithm} \\ \hline
1:    &   Choose a small set of  initial explored points $P_{exp} = \{{\bf x}_1, {\bf x}_2, .... {\bf x}_{N_{min}}\}$ using a maximin \\
      &    Latin hypercube design, {and evaluate $f({\bf x}_i)$ on $P_{exp}$}\\
2:    &   {\bf for } $N=N_{\min},...,N_{\max}$  {\bf do}                            \\
3:    &    \quad Construct a Bayesian surrogate model 
$\sum_{i=1}^{N} \beta_{i} r({\bf x}; \bfmu_i, s_i)+  \epsilon({\bf x})$  \\
      &     \quad  as in Section 3.1 based on $\{({\bf x}_i, f({\bf x}_i)), i=1,..., N\}$   \\
4:    &   \quad  Calculate the SEI in (\ref{eq: EI}), and \\ 
    & \quad select a new explored point ${\bf x}_{N+1}$ via selection criterion in  
    (\ref{eq:arg_max_sn}) over $V \setminus P_{exp}$ \\
6:  & \quad Update  $P_{exp} = P_{exp} \cup  {\bf x}_{N+1}$ {and evaluate $f({\bf x}_{N+1})$} \\
7:   &  {\bf end for}\\
8:   &   {\bf Return} the current best optimal point, $\hat{{\bf x}}_{\text{opt}} =\arg\max_{\bf x \in P_{exp}} f({\bf x})$ and \\
    & \quad the corresponding function value, $f(\hat{{\bf x}}_{\text{opt}})$.\\
 \hline
\end{tabular}
\end{table}
Note that the proposed {BaRBF} can be flexibly used in different
scenarios. For example, when the number of available function
evaluations is small to moderate, there may not be enough
observations to estimate all the parameters. In this case, we only
need to update some part of 
RBF parameters, say the scale parameter $s$ by setting all the scale
parameters $s_i \equiv s,$ $(i=1,...,N),$ and need not update the
$\mu_i$ parameters. The choice of whether to update all parameters
or part of them can be decided based on the magnitude of the model
residuals at the initial stage. If updating all parameters leads to
relative large model residuals, then we can fix certain parameters
instead. The formulas of the posterior distribution in (\ref{eq:
full})-(\ref{eq: proposed3}) need some minor changes accordingly if
certain RBF parameters are fixed. For the above example, one only
needs to set $s_i$ in eq. (\ref{eq: full}) to be the same $s$, and
update only one $s$ in (\ref{eq: proposed3}), and does not need to
update the $\mu_i$'s in (\ref{eq:post_mu}) and (\ref{eq: proposed}).

For the remaining part of this section, we will compare our {BaRBF}
with the Global metric stochastic RBF ({G-MSRBF}) algorithm proposed  by
\cite{regis2007stochastic}  from a theoretical perspective. The {G-MSRBF}
method will be regarded as the baseline method from now on. First we
give a brief review. The {G-MSRBF} employs a surrogate model $s_N({\bf x})$ using RBFs,
\begin{eqnarray} \label{eq:RBF_sN}
s_N({\bf x}) =  \sum_{i=1}^{N} \lambda_i r({\bf x}; {\bf x}_i, s).
\end{eqnarray}
The RBF parameters in (\ref{eq:RBF_sN}) are pre-specified, i.e.,
the RBF centers are set at the explored points ${\bf x}_i$, and $s$
is pre-calculated at the initial stage of optimization. The model
coefficients $\lambda_i$ in (\ref{eq:RBF_sN}) are estimated by
solving a deterministic linear system of equation $\Phi \lambda =
F,$ where $\Phi_{ij} = r({\bf x}_i; {\bf x}_j, s),$ $F = (f({\bf
x}_1), ..., f({\bf x}_N))^T.$ And their point selection criterion
\begin{eqnarray}
W_N({\bf x}) =  (1-\omega_N^G)  V_N^R({\bf x}) + \omega_N^G V_N^D({\bf x}).
\end{eqnarray}
is a weighted average of the scaled response prediction $V_N^R({\bf x})$ with
\begin{eqnarray}
V_N^R({\bf x})=\begin{cases}
                (s_N({\bf x}) - s_N^{\min})/(s_N^{\max} - s_N^{\min}) \text{ for } s_N^{\max} \neq s_N^{\min}, \\
              1 \text{ o.w,}
            \end{cases}
\end{eqnarray}
and the maximin distance criterion $ V_N^D({\bf x})$ with
\begin{eqnarray}
V_N^D({\bf x}) = (d_N({\bf x}) - d_N^{\min})/(d_N^{\max} - d_N^{\min}),
\end{eqnarray}
where $s_N^{\max} = \max\{s_N({\bf x})\},$ $s_N^{\min} =
\min\{s_N({\bf x})\},$ $d_N({\bf x}) = \min_{1\le i \le N} ||{\bf
x}-{\bf x}_i||^2,$ $d_N^{\min} = \min d_N({\bf x}),$ $d_N^{\max} =
\max d_N({\bf x}).$ The $\omega_N^G$ can take values in $\{1, 0.8,
0.6, 0.4, 0.2\}$ periodically. For example, if at time $N=20,$
$\omega_N^G=0.8,$ then at the next time $N=21,$ $\omega_N^G=0.6.$
Then a new point ${\bf x}_{N+1}$ is selected to maximize $W_N({\bf
x})$, and {finally the global maximizer is also  estimated by ${\bf
\hat{x}}_{opt}= \arg \max_i f(x_i)$.}

Although both methods use RBFs, there are two main differences.
First, the surrogate model is different.
{BaRBF} uses a Bayesian surrogate model that provides not only
predictions but also its uncertainties, while the {G-MSRBF} utilizes a
deterministic surrogate model that only provides predictions.
Because our proposed surrogate model is similar to ridge regression,
the approximation of response is more robust and smooth compared to
the interpolation surrogate model in {G-MSRBF}. The second difference
lies in the choice of the selection criterion for new explored
points. In our method, we utilize the expected improvement criterion $E(\max\{y -
f_{\max}, 0\}),$ which can be regarded as a {\it soft-thresholding}
version of $E(y)$. As previously discussed, thresholding the
prediction makes it easier to identify global optima. In addition, under the Gaussian
prediction, EI criterion contains the measure of the prediction improvement and the model uncertainty. Here the part of the prediction improvement can be treated as exploitation and the uncertainty part is used to explore the search space. In {G-MSRBF}, the global exploration is based on the maximin distance criterion, $V_N^D({\bf x})$, and the $V_N^R({\bf x})$ is used for local refining. The weighted average of these two criteria is adopted with pre-specific weight pattern. Simulation studies will be presented in
Sections \ref{sec:simulation} and 5 to further understand and compare the
empirical performance of the two methods.



\section{Simulation Study} \label{sec:simulation}

To assess the performance of {BaRBF}, we compare it with {G-MSRBF}, which is
regarded as the baseline method. In {G-MSRBF}, the candidate set for the next  point selection is chosen in each iteration. To make a fair
comparison, the candidate points are fixed as a pre-specified grid, $\chi$, in the experimental region and both methods will be implemented over the same grid.


In addition to the {G-MSRBF}, we also consider another global optimization approach based on Gaussian process surrogate model for comparisons.
\cite{jones1998efficient} proposed the efficient global optimization (EGO) approach by using the Gaussian process for surrogate construction.
EGO starts from an initial point set. After evaluating the response values of the initial design points, a numerical optimization approach, like meta-heuristic algorithm, is used to obtain the MLE of the parameters in the Gaussian process and then the corresponding surrogate model is obtained.
Since the Gaussian process prediction follows a normal distribution, the EI criterion in Eq. (\ref{EI_n}) is used to identify the next explored point over the feasible point set, $\chi \setminus P_{exp}$, and then the surrogate model is updated.
Iterate these two steps until a stopping criterion is met. Usually the stopping criterion can be the number of explored points or the maximum value of the EI criterion over the unexplored points.


First we start with a 2D global optimization example, whose objective function is quit smooth. Then we consider another 2D objective function which is not smooth and has multiple global optima.

\subsection{2D Branin function} \label{sec_braini}

\begin{figure}
\centering 
 \includegraphics[width=3.5in]{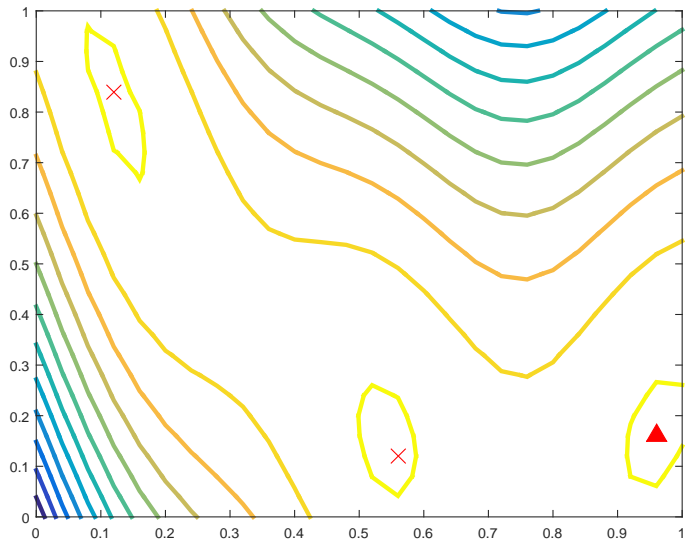}\\
 \protect\caption{The contour plot of the Branin function on $[0, 1]^2$ with grid size 0.04. The red triangle represents the global optimum and two red crosses denote the other two local optima.}
\label{fig:true_brainin}
\end{figure}

We consider the standard 2D test function ``Branin function'', which
has been widely used in the global optimization literature, e.g.
\cite{jones1998efficient}. The scaled version of ``Branin function''
we use here is defined as follows,
\begin{eqnarray} \label{eq:Brainin}
f({\bf x}) = \frac{-1}{51.95}[(\bar{x}_2 - \frac{5.1{\bar x}_1^2}{4\pi^2} + \frac{5\bar{x}_1}{\pi} - 6)^2 + (10-\frac{10}{8\pi})\cos(\bar{x}_1) - 44.81],
\end{eqnarray}
where $\bar{x}_1 =   15 x_1 - 5, \bar{x}_2 = 15x_2,$ and $x_1 \in
[0,1], x_2 \in [0,1].$ To simplify our code, we further restrict
this function on {the} evenly spaced grid $\chi = [0, 0.04, \dots,
1]^2$. The contour plot of the Branin function {over the
pre-specified grid points} is given in Figure
\ref{fig:true_brainin}, where there will be two local maxima and one
global maximum on $[0.96, 0.16]$ with the maximum value 1.0473. In
{BaRBF}, we first measure the prediction uncertainties for all grid
points in $\chi$ and then identify the next explored point via
(\ref{eq:arg_max_sn}) from the set $\chi \setminus P_{exp}$.

The objective is to find ${\bf x}$ that maximizes $f(\mathbf{x})$ in
(\ref{eq:Brainin}) with as few evaluations as possible. At each
iteration of the algorithm, {the current optimal point, ${\bf
\hat{x}}_{opt}$, and its function value $f({\bf \hat{x}}_{opt})$ are
recorded together with all the explored points.} We randomly choose
a small set of  $N_{min} (= 16)$ initial explored points using a
maximin Latin hypercube design \citep{santner2013design}. All three methods start with the same  set of ${\bf x}_i$'s. Each time the
surrogate model is updated by incorporating the $f$ value of a new
explored point. {Then} we calculate and update the ${\bf
\hat{x}}_{opt}$ value. For each algorithm, new explored points are
selected and evaluated sequentially until the total number of
explored points reaches $N_{max} (= N_{min} + 30) = 46.$ This
process is repeated 60 times, and the average performances are
reported and compared for the two methods.

For fair comparison among {BaRBF} and {G-MSRBF}, we set the initial sampler of the RBF
parameters in {BaRBF} to be the same as the fixed RBF parameters in
{G-MSRBF}. Specifically, we use Algorithm 1 in
\cite{fasshauer2007choosing} to select an optimal value of $s$ in
{G-MSRBF} that minimizes a cost function that collects the errors for a
sequence of partial fits to the data. The center parameters
$\bfmu_i$'s are set as the explored points ${\bf x}_i$'s.

{From many simulation trials,} we found out that, for the Branin
test function, updating all parameters in {BaRBF} will lead to
relatively large model residuals that do not converge. This might be
caused by the small number of function evaluations. Thus we only
update one scale parameter $s$ with all $s_i \equiv s$ and fix the
center parameter $\mu_i$'s {at} the explored  points. We iterate
the MCMC 10,000 times. Also, we discard the first 40\% of the
samples, and take 1 out of every 5 samples in the remaining 60\% of
the samples, in order to obtain stable and less correlated posterior
samples for model fitting. In order to implement {BaRBF}, two important
tuning parameters need to be pre-specified. The first one is the
value of $C$ in the coefficient prior. Here we set $C = 25$. Since
the weighted selection criterion is adopted to select the next
explored point, the parameter $d$ in the weight $\omega_N$ is fixed
as 5 after we have tested several different possibilities.

\begin{figure}[t]
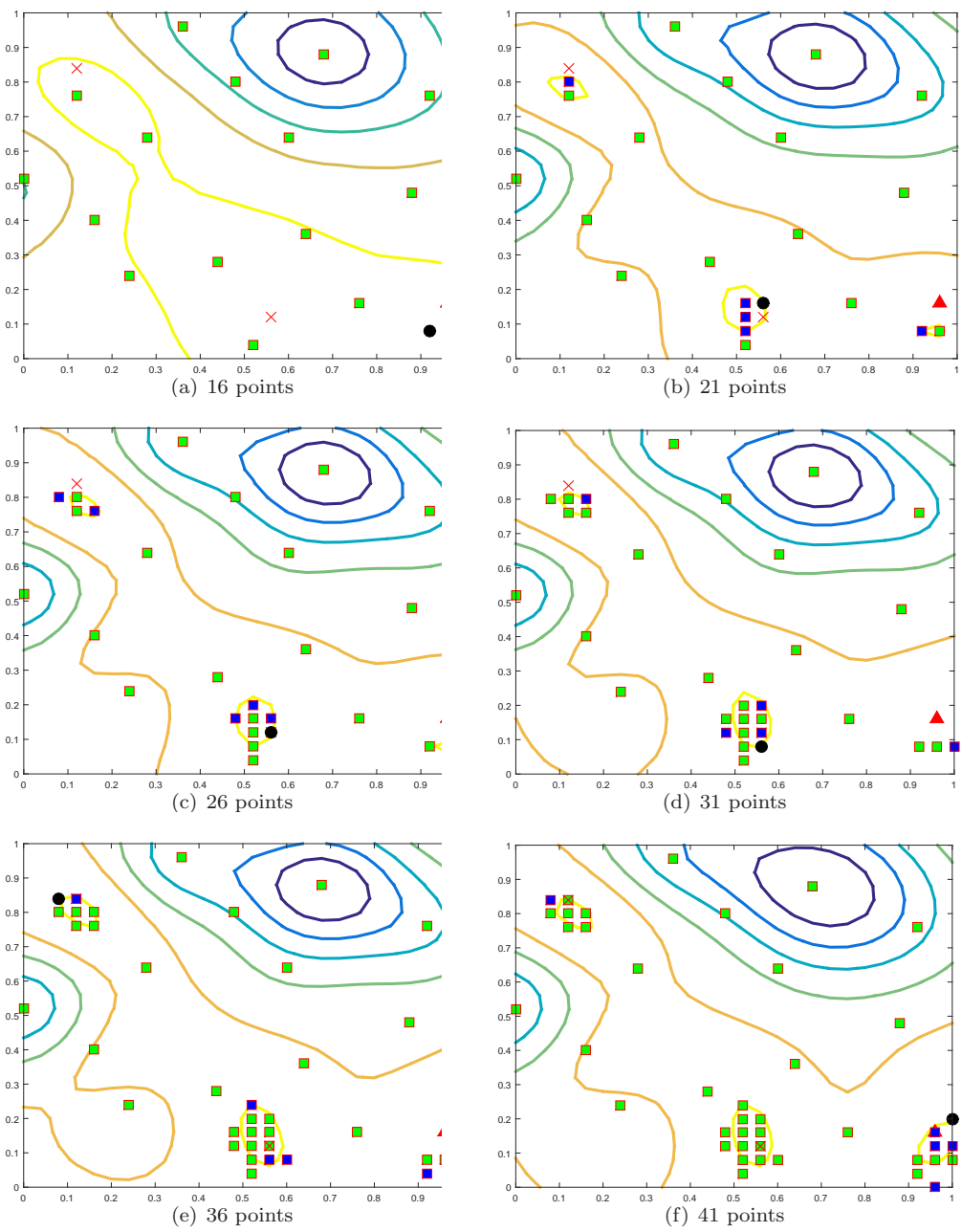

\begin{center}
\begin{tabular}{cc}
  \subfigure[16 points ]{\includegraphics[width=2.5in]{is16pts_SEI.eps}}  &
  \subfigure[21 points ]{\includegraphics[width=2.5in]{ps21pts_SEI.eps}} \\
  \subfigure[26 points ]{\includegraphics[width=2.5in]{ps26pts_SEI.eps}} &
  \subfigure[31 points ]{\includegraphics[width=2.5in]{ps31pts_SEI.eps}}  \\
  \subfigure[36 points ]{\includegraphics[width=2.5in]{ps36pts_SEI.eps}}  &
  \subfigure[41 points ]{\includegraphics[width=2.5in]{ps41pts_SEI.eps}}  \\
\end{tabular}
\end{center}
\protect\caption{The contours of surrogate model using {BaRBF}. Each of
the six plots corresponds to a surrogate model with N = 16, 21, 26,
31, 36, 41 respectively. (Explanation of symbols is given in the
text.) } \label{fig:1_brainin}
\end{figure}

In this subsection, we first illustrate the proposed {BaRBF} with one
particular simulation sample. 
{Figure \ref{fig:1_brainin} plots the contours of the surrogate
model in {BaRBF} and the locations of the explored points using {BaRBF}
with $N=16,$ 21, 26, 31, 36, 41 for one simulation sample. Figure
\ref{fig:1_brainin}(a) shows the initial status of {BaRBF}. The initial
design is a 16-run maximin Latin hypercube indicated by 16 green
squares. The next explored point chosen by the selection criterion,
i.e. the 17th point, is indicated by the black circle in the lower
right corner. In Figure \ref{fig:1_brainin}(b), the five additional
points (17th to 21st) are indicated by the five blue squares.
These five points are divided into three sets, one closer to the global maximum, the other closer to the other two  local maximums. As in Figure
\ref{fig:1_brainin}(a), the next explored point, i.e., the 22nd point, is indicated by the black circle. In Figure \ref{fig:1_brainin}(c), all the 21 points from Figure \ref{fig:1_brainin}(b) are indicated by green squares, the additional five points by blue squares, and the next explored point by black circle. Then the same symbols are used in Figure
\ref{fig:1_brainin}(d) to (f) to demonstrate the progression of
points for $N = 31, 36$ and 41. 
Amazingly, except the initial design points, all the explored points are located closer to the three maxima, none for exploring bad regions. Finally the global maximum is identified in
point 39 as shown by the black square in Figure
\ref{fig:1_brainin}(f). In summary these contour plots show that
{BaRBF} efficiently explores the experimental space and quickly
approaches the optimal points.}

\begin{table}[t] \caption{Summary of optimal values
obtained by {BaRBF}, {G-MSRBF} and EGO with 60 replications in the 2-dimensional
experiment with Branin function. The run size of the initial design is 16, and the
optimal value of the Branin function is 1.0473.} \label{2-dim}
{\small
\begin{center}
\begin{tabular}{|c|l|l|l|l|l|l|l|l|}
  \hline
  Approach & 5\%   & Q1 & Median & Q3 & 95\%  & Mean & std & Frequencies with \\
           & Quantile&    &        &    & Quantile&      &     &true optimal values \\ \hline
  {BaRBF}(SEI)& 1.0397 & 1.0397&   1.0464 &  1.0473 &  1.0473 &  1.0448 & 0.0033& 29/60\\ \hline
  {G-MSRBF} & 1.0176& 1.0438 & 1.0464 &1.0473 & 1.0473 &1.0425 & 0.0152 &20/60  \\ \hline
  EGO & 1.0473 & 1.0473 &  1.0473 & 1.0473 & 1.0473 & 1.0473& 0.000 & 60/60 \\ \hline
\end{tabular}
\end{center}}
\end{table}

\begin{figure}
\centering
\includegraphics[width=5.5in]{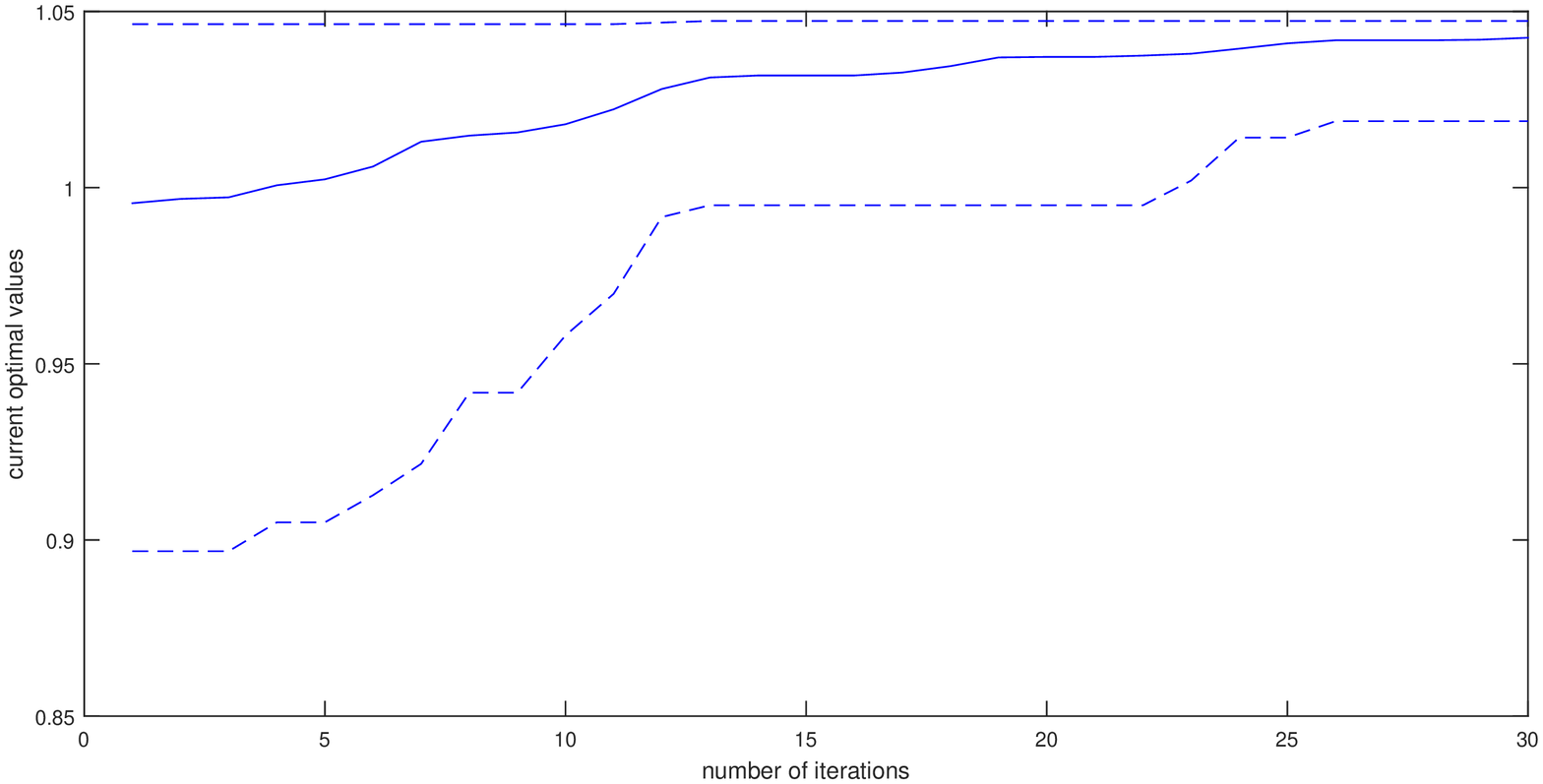}\\
 \bigskip
 \includegraphics[width=5.5in]{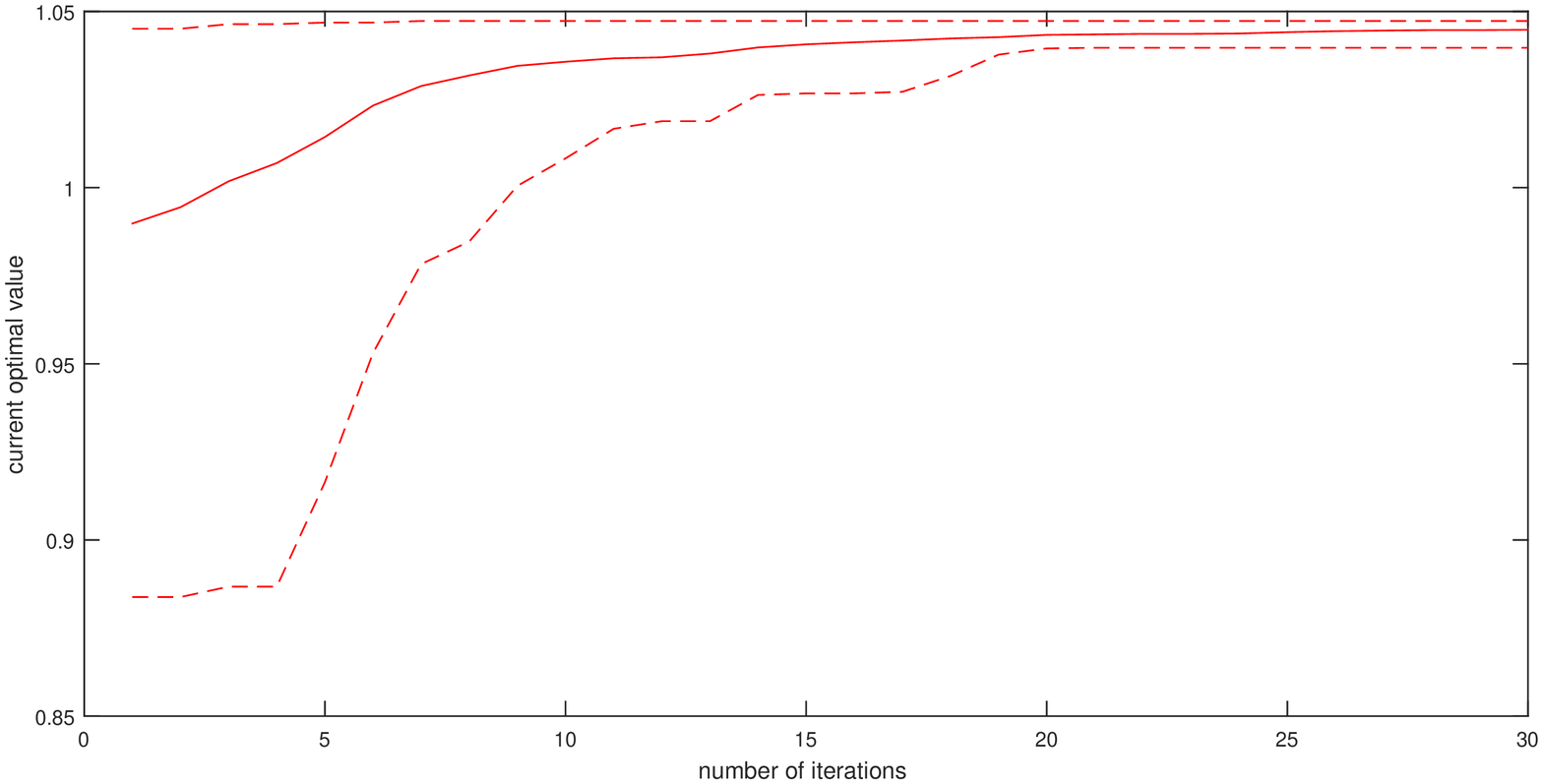}\\
 \protect\caption{The mean value (solid line) and the 5\% and 95\% quantiles (dashed line) of current optimal
values based on 60 replications for the example of Branin function. Upper panel: {G-MSRBF}, lower panel:
{BaRBF}.} \label{fig:multi_brainin}
\end{figure}

We report the performance of {BaRBF} and {G-MSRBF} based on 60 replications
by randomly generating the initial LHD designs. The purpose is to
see whether {BaRBF} provides a more efficient search path to identify
the global maximum compared with {G-MSRBF}, for the same number of
function evaluations. The numerical results are summarized in Table
\ref{2-dim}. First {BaRBF} has the higher mean value, 1.0443, than that
of {G-MSRBF}, 1.0425, and is more stable because of its smaller standard
deviation. Based on the sample quantiles of the optimal values
identified by both approaches, there is a detectable difference at
the 5\% quantile values. We found out that for several cases, the
best points identified by {G-MSRBF} are not close to the true optimal
point {and after checking the corresponding search processes, {G-MSRBF}
did not efficiently explore the experimental space by properly
choosing the next points.} On the other hand, {G-MSRBF} performs better
than {BaRBF} for the first quartile Q1. In addition, among the 60
replications, {BaRBF} can identify the true optimal values 26 times,
which is higher than 20 times for {G-MSRBF}. Overall
{BaRBF} has a better performance. 
Here we plot in Figure \ref{fig:multi_brainin} the mean value as
well as the 5\% and 95\% quantile curves of the current optimal
values with respect to the number of iterations for both methods.
The mean curves in the two plots are very similar. More meaningful
is the comparison of the two 5\% quantile curves. For {G-MSRBF}, the
curve moves up quickly until $N$ =13; then it gets stuck (flat)
until about $N$ = 23. By comparison, the 5\% quantile curve for
{BaRBF} moves up quickly until $N$ = 18. By then, the band between the
upper and lower quantile curves is very narrow and continues to
shrink. The corresponding band for {G-MSRBF} does not shrink even  to the
end ($N$ = 30). In fact it remains very wide when $N$ = 26. This
figure gives a more informative comparison than the numerical
results in Table \ref{2-dim}. It clearly shows the better
performance of {BaRBF} over {G-MSRBF}.

When we compare the results with EGO, it seems that EGO works perfectly in this Branin example because EGO can quickly identify the global maximum point in each replications. The possible reason should be that since the Branin function can be treated as a smooth function, it can be fitted quit well by the Gaussian process model and thus the EI criterion in EGO can rapidly guide the search process to the target point. For our {BaRBF}, we do have a error assumption in the surrogate model and the model fitting may not be as well as Gaussian process model. In addition, SEI is computed as the sample expectation of the improvement function without any distributed assumption. Thus if the Gaussian assumption is satisfied, it is not surprised that EI can be more efficient in getting the global optimal point. In fact, when we monitor the search process of {BaRBF}, sometimes it may stay in a local area for a while. This should be related to that the exploration effect of the SEI criterion does not function well.

\subsection{2D Ronkkonen function}

\begin{figure}[t]
\centering 
 \includegraphics[width=5in]{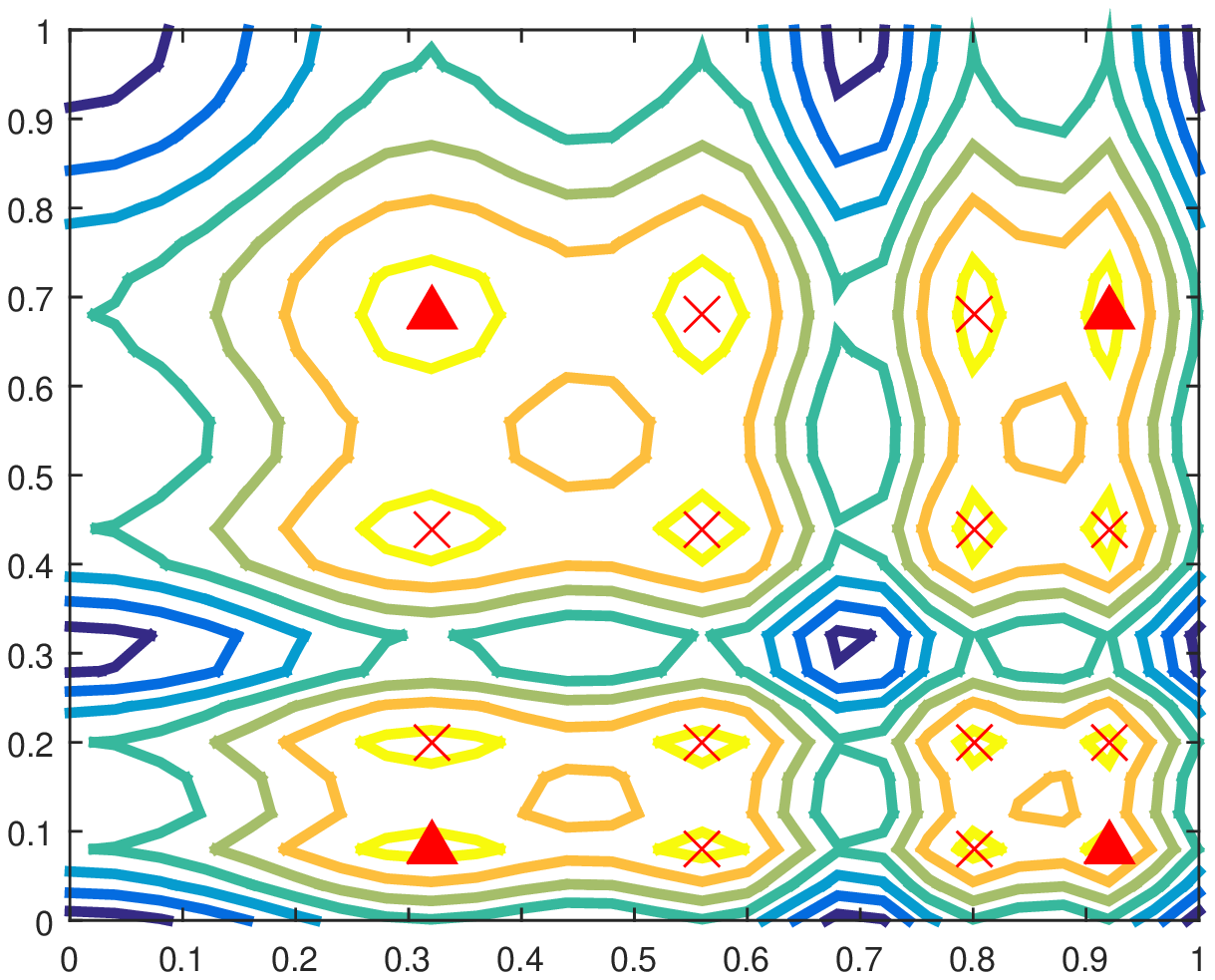}\\
 \protect\caption{{The contour plot of the Ronkkonen function on $[0, 1]^2$ with grid size 0.04. The red triangle represents the global optimum and 12 red crosses denote the other two local optima.}}
\label{fig:true_ronkkone}
\end{figure}

In addition, we consider another 2D objective function in \cite{ronkkonen2011}, i.e.,
\begin{eqnarray}
f(x_1, x_2) = -\frac{1}{4}\sum_{i= 1}^2 [\cos(4\pi w_i) + 0.8 \cos(8 \pi w_i)],
\end{eqnarray}
where $w_i = \sum_{j=0}^{n_i} {{n_i} \choose {j}} P_{ij} (1-x_i)^{n_i-j} x_{i}^{j}$ for $i=1, 2$, $n_1 = n_2 = 4$, and $P_1 = (0, 0.1, 0.2, 0.5, 1),$ $P_2 = (0, 0.5, 0.8, 0.9, 1)$ and the experimental region is $[0,1]^2$.
This objective function has been used as a test function in \cite{chipman2012}.
Here we also restrict the function on the evenly spaced grid $\chi = [0, 0.04, \dots, 1]^2$. The contour plot of this Ronkkonen function over the pre-specified grid points is given in Figure \ref{fig:true_ronkkone}, where there are 12 local maximums and 4 global maximal points with the maximum value, 0.4777. Because of its multiple local and global optimal points, this Ronkkonen function is not as smooth as the Branin function.

{In this example, the initial point sets are the same as those in Section 4.1, and the total number of explored points is set as $N_{max} = 16 + 30 = 46$, i.e., based on 16 initial points, the search algorithm iterates 30 times by sequentially adding 30 points. Then the best value among the 46 explored points is reported. Here the goal is to identify one of the four global maximum points. The average performances of {BaRBF}, {G-MSRBF} and EGO over 60 replications are summarized in Table \ref{2-dim-Ro}.}


\begin{table}[t] \caption{Summary of optimal values
obtained by {BaRBF}, {G-MSRBF} and EGO with 60 replications of the Ronkkonen function. (The run size of the initial design is 16, and the
optimal value of the Ronkkonen function is 0.4777.)} \label{2-dim-Ro}
{\small
\begin{center}
\begin{tabular}{|c|l|l|l|l|l|l|l|l|}
  \hline
  Approach & 5\%   & Q1 & Median & Q3 & 95\%  & Mean & std & Frequencies with \\
           & Quantile&    &        &    & Quantile&      &     &true optimal values \\ \hline
  {BaRBF}(SEI) &0.4766 &   0.4775&   0.4777 &  0.4777 &  0.4777 & 0.4775 & 4.6850e-4 &43/60 \\ \hline
  {G-MSRBF}   &0.3635  & 0.4407	& 0.4766 & 0.4775 & 0.4777	& 0.4529 & 0.0363 & 11/60\\ \hline
  EGO & 0.3922	& 0.4405&0.4766	&0.4777&	0.4777 &0.4526 & 0.0344 &18/60  \\ \hline
\end{tabular}
\end{center}}
\end{table}

\begin{figure}[t]
\centering
\includegraphics[width=5.5in]{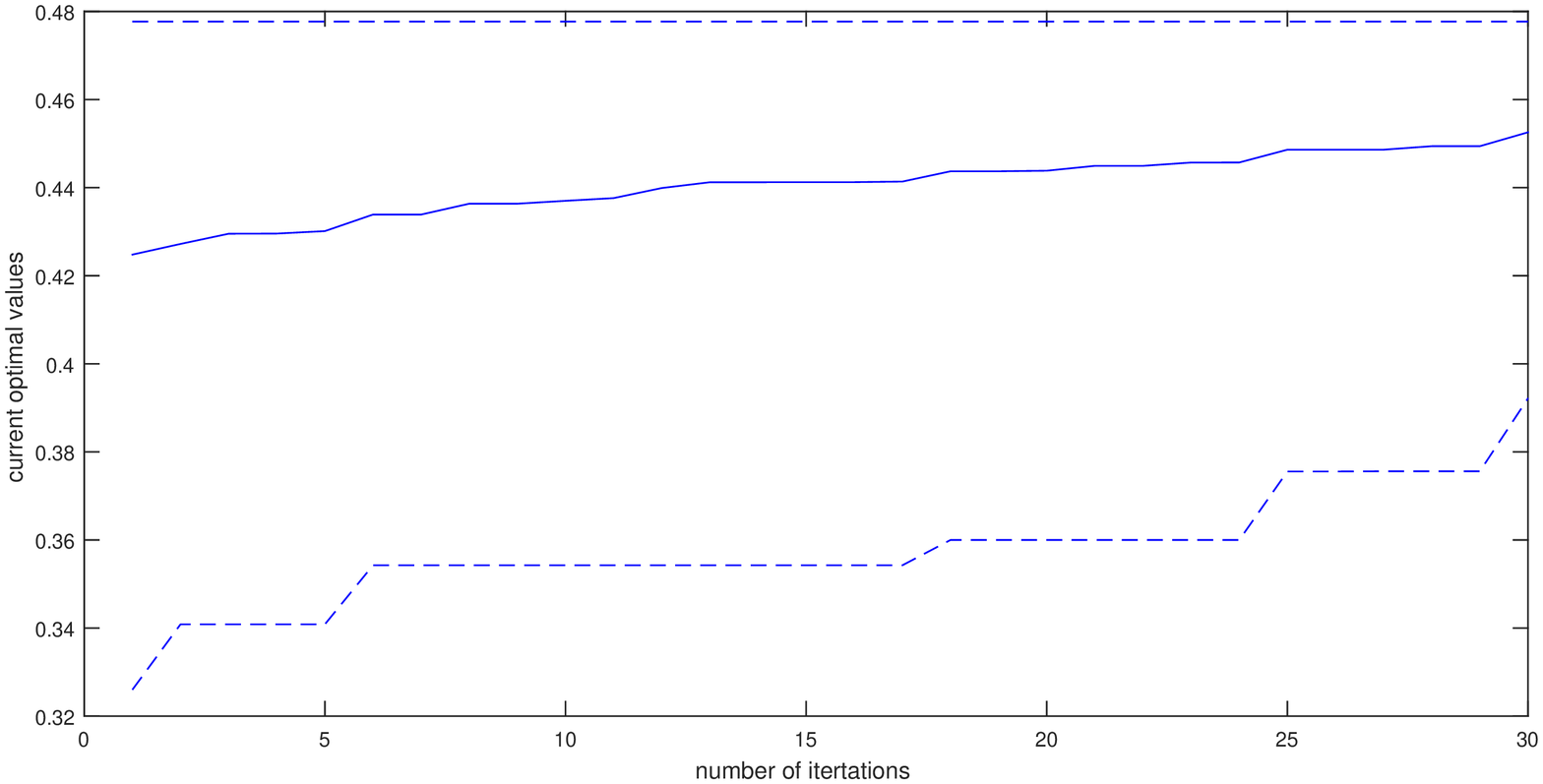}\\
 \bigskip
 \includegraphics[width=5.5in]{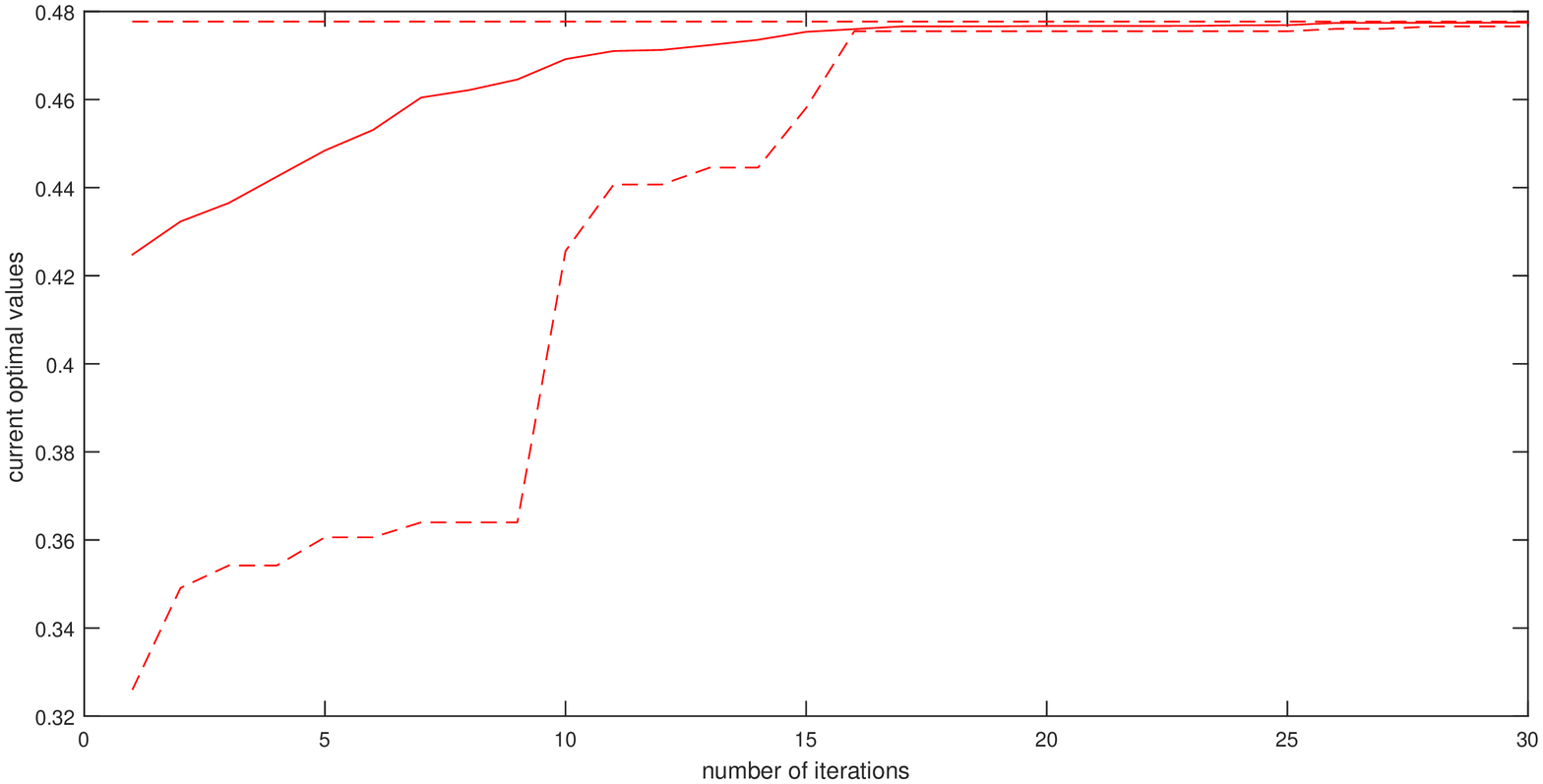}\\
 \protect\caption{{The mean value (solid line) and the 5\% and 95\% quantiles (dashed line) of current optimal values based on 60 replications, 
 Ronkkonen function. Upper panel: EGO, lower panel:
{BaRBF}.}} \label{fig:multi_ronkkone}
\end{figure}

{First, {G-MSRBF} performs worst in terms of the frequencies of reaching  one of the
four global maximum points (see the last column of Table \ref{2-dim-Ro}).
Then we focus on comparing the performance between {BaRBF} and EGO.
From Table \ref{2-dim-Ro}, the mean of the best function value found by {BaRBF} is 0.4775 with standard deviation 4.6850e-4, while the corresponding values for EGO are 0.4526 and 0.0344 respectively. We also compute the sample quantiles of the best values found by both methods. Table \ref{2-dim-Ro} shows that {BaRBF} touches the global maximum at the $50\%$ sample quantile, while EGO reaches 0.4777 at the $75\%$ quantile. In addition, for the {BaRBF}, the frequency of reaching a global optimum is 43/60, while the frequency for the EGO is 18/60.
The mean value of 60 replicates, and the $5\%$ and $95\%$ quantile curves of the current optimal values with respect to the number of iterations for EGO and {BaRBF} are shown in Figure \ref{fig:multi_ronkkone}.
The mean curve of the {BaRBF} moves up quickly to the global optimal value, while the curve for EGO moves up more slowly.
In addition, the $5\%$ quantile curve for EGO does not get much improvement before adding $25$ explored points, i.e., $N = 16+25 = 41$, while the same curve for {BaRBF} moves up quickly and gets close to the global optimal value after adding 15 points, i.e., $N = 16 + 15 = 31$. Another attractive feature for {BaRBF} is the extremely low standard deviation (see the std column), which may suggest that the {BaRBF} can perform stably over repeated implementations.
In summary, the {BaRBF} outperforms the EGO in this example.
}

{
\cite{chipman2012} have pointed that when the test function has multiple global optima, the EI criterion might lead the search process to jump out of the neighborhood of one global optimum to another one and cannot stick around one global optimum.
This may be explained by the fact that the EI criterion tries to minimize the prediction uncertainties among different global optima. The $5\%$ quantile curve for EGO in Figure \ref{fig:multi_ronkkone} seems to support this point.
For the {BaRBF}, by using SEI, it can quickly locate one neighborhood of a global maximum and then identify the best value. 
In addition, since this Ronkkonen function is not as smooth as the Branin function, the normality assumption and the interpolation property of the Gaussian process may not give advantages for the surrogate construction.
For the {BaRBF}, the surrogate model consists of  additive radial basis functions and their parameters are simultaneously adjusted via the proposed Bayesian approach. Thus our surrogate construction approach may be more advantageous for non-smooth test functions.
}

\section{Simulation Studies for Three and Four Dimensions}
\label{sec:higher_dim}

In addition to these two different 2D test functions, we consider 3D and 4D examples to illustrate that the proposed {BaRBF} can deal with higher dimension cases and also compare its performances with EGO and {G-MSRBF}.

\subsection{3D Ronkkonen Function}

The Ronkkonen function defined in \cite{ronkkonen2011} is quite general and can be extended to different dimensions.
Here we consider a 3D Ronkkonen function which is generalized from the 2D function in Section 4.2 and is shown below,
\begin{eqnarray}
f(x_1, x_2, x_3) = -\frac{1}{4}\sum_{i= 1}^3 [\cos(4\pi w_i) + 0.8 \cos(8 \pi w_i)],
\end{eqnarray}
where $w_i = \sum_{j=0}^{n_i} {{n_i} \choose {j}} P_{ij} (1-x_i)^{n_i-j} x_{i}^{j}$ for $i=1, 2, 3$, $n_1 = n_2 = 4$, and $P_1 = (0, 0.1, 0.2, 0.5, 1)$; $P_2 = (0, 0.5, 0.8, 0.9, 1)$; $P_3 = (0, 0.6, 0.7, 0.9, 1)$. The experimental region considered here is $[0,1]^3$. According to \cite{ronkkonen2011}, there are $5^3$ local maximum points and 27 of them are the global maximum points.

We divide the experimental region into
$(26)^3$ grid points by setting the grid size as 0.04.
For this grid, there is only one global maximum point $(0.3200, 0.6800, 0.4400)$ with the function value 0.3584.
We start by choosing the initial maximin LHD with 50 points for the 3-dimensional case.
After the initial stage, we run {BaRBF} for 50 iterations. To implement {BaRBF}, we follow the same RBF set-up in Section 4.1. That is we set the scale
parameter $s$ with all $s_i \equiv s$ and fix the center parameter
$\mu_i$'s at the explored points. For the tuning variable $C$, 
we fix $C$ as 15. The performance of the {BaRBF} is measured by the maximum function value identified within 100 explored points and is summarized based on 20 replications by independently re-generating the initial design points.
For the comparison purpose, we also implement EGO and {G-MSRBF}.

Table \ref{3-dim} is a summary of the maximum values
obtained by the three approaches.
{G-MSRBF} performs worst in this case because {G-MSRBF} cannot identify any true global maximum value within 20 replications.
Between {BaRBF} and EGO, {BaRBF} has better performance in the average maximum function values and the frequency of reaching the global maximum point, but both approaches share similar values in the first and third quartiles, Q1 and Q3, and the median value.
Because there are few replications, the maximum value identified by EGO is less than 0.34.
A possible reason should be similar to what was stated in Section 4.2, namely, the Ronkkonen function is not a smooth function and EGO may jump around different local modes. Finally, the std value for {BaRBF} is much lower than that for the others. This is similar to what we observe in Table \ref{2-dim-Ro} for the 2D case and has a similar implication on the stability of the {BaRBF}.

\begin{table}[t] \caption{Summary of maximum values
obtained by {BaRBF}, {G-MSRBF} and EGO with 20 replications in the 3-dimensional case. The optimal value of the 3-dimensional Ronkkonen function is
0.3584.} \label{3-dim}
\begin{center}
\begin{tabular}{|c|l|l|l|l|l|l|}
  \hline
  Approach & Q1 & Median & Q3 & Mean & std & Frequencies with \\
           &    &        &    &      &     &true optimal values \\ \hline
  {BaRBF}(SEI) & 0.3578 & 0.3580 & 0.3584& 0.3581&	3.3973e-04	& 9/20 \\ \hline
  {G-MSRBF} & 0.3203	& 0.3362 & 0.3576 & 0.3334 & 0.0234&	0/20 \\ \hline
  {EGO} &0.3579 & 0.3580 & 0.3583 & 0.3566 & 0.0048 & 5/20 \\ \hline
\end{tabular}
\end{center}
\end{table}

\subsection{4D Hartmann Function}

In this section, we study the performance for the 4-dimensional Hartmann function. 
This function has been
studied in \cite{picheny2013} and is defined as
  \begin{eqnarray}
  f(x_1, x_2, x_3, x_4) = -\frac{1}{0.839}[1.1 - \sum_{i=1}^{4} \alpha_i
  \exp(-\sum_{j=1}^4  A_{ij} (x_j - P_{ij})^2)],
  \end{eqnarray}
  where $\alpha = (\alpha_i) = (1.0, 1.2, 3.0, 3.2)$; $A = (A_{ij})
  = \left(
      \begin{array}{cccc}
        10 & 3 & 17 & 3.5 \\
        0.05 & 10 & 17 & 0.1 \\
        3 & 3.5 & 1.7 & 10 \\
        17 & 8 & 0.05 & 10 \\
      \end{array}
    \right)$ and $P = (P_{ij}) = 10^{-4} \left(
                                           \begin{array}{cccc}
                                             1312 & 1696 & 5569 & 124 \\
                                             2329 & 4135 & 8307 & 3736 \\
                                             2348 & 1451 & 3522 & 2883 \\
                                             4047 & 8828 & 8732 & 5743 \\
                                           \end{array}
                                         \right).$
There are multiple maxima for this function over the experimental
region $[0,1]^4$. Here we divide the experimental region int
$(21)^4$ grid points by setting the grid size as 0.05. The maximum
function value over these points is 3.1218.

The proposed method, {BaRBF}, with SEI selection criterion is used to identify the true optimal
point over the pre-specified grid for this 4D function.
For the parameters in radial basis function, the scale parameter $s$ is set as $s_i \equiv s$ and the center parameters,
$\mu_i$'s, are fixed at the explored points. In addition, the variance, $C$, in the mixture
normal priors of the coefficients is fixed as 10. 

{We start by choosing the initial maximin LHD with 
50 points for the 4-dimensional case.
After the initial stage, we run {BaRBF} for 50 iterations. Thus there
are in total 100 explored points in this example. To track the performance of the proposed method,
we record the maximum function values in each iteration, i.e.,
$\max_{\mathbf{x} \in \mathcal{P}_{exp}} f(\mathbf{x})$. That was
repeated 20 times by randomly generating the initial maximin LHD. We
repeat the same procedure for {G-MSRBF} and EGO. A summary of the maximum values
obtained by the three approaches are shown in Tables
\ref{4-dim}. For this 4-dimensional Hartmann function, {BaRBF} outperforms {G-MSRBF} in terms of the frequency, the mean of the optimal
values and a smaller standard deviation, and as shown in Table \ref{4-dim}, for {BaRBF}, the middle 50\% of values between Q1 and Q3 is extremely tiny and smaller than that for {G-MSRBF}. However, EGO has the best performance in this case. We think it should be related to the target function because this Hartmann function should be a smooth function and thus it is fever to the EGO approach.
}



\begin{table}[t]
\caption{Summary of maximum values obtained by {BaRBF}, {G-MSRBF} and EGO
with 20 replications in the 4-dimensional case. The optimal value of
the 4-dimensional Hartmann function  is 3.1218.} \label{4-dim}
\begin{center}
\begin{tabular}{|c|l|l|l|l|l|l|}
  \hline
  Approach & Q1 & Median & Q3 & Mean & std & Frequencies with \\
           &    &        &    &      &     &true optimal values \\ \hline
  {{BaRBF}(SEI)} & 3.1119&3.1218& 3.1218&3.0936 &  0.0746 & 15/20 \\ \hline
  {G-MSRBF} & 3.0948&3.1218& 3.1218&3.0903 & 0.0972 & 13/20 \\ \hline
  {EGO} & 3.1218  &  3.1218 & 3.1218 &  3.1099 & 0.0531& 19/20 \\ \hline
\end{tabular}
\end{center}
\end{table}

\section{Discussions}

In this session, several issues are studied. First we modify the proposed algorithm by adding a step to force {the} search process to jump out from a local optimal area. Then we study the effects of grid size on the performance of the proposed method. {Finally, in order to reduce the computational burden for high-dimensional optimization problems, a grid-free} version of the proposed method is introduced and {a} higher dimension example is illustrated.

\subsection{{A Modified Version of {BaRBF}}}\label{sec:M-aBRF}

{From tracing the search process of the {BaRBF} in the Branin function example, we found out that sometimes the {BaRBF} gets stuck in a local area and cannot leave the area for a while. 
To overcome this potential weakness, we have the following modification.
To jump out of this local area, we add an additional step, called the {\it escape step}, by monitoring the search process of the current best value.
That is, we record the number of consecutive non-improvement iterations, i.e., iterations for which the current best function value cannot get improved from the new explored point.
Denote this number by $C_{non}$. Once $C_{non}$ exceeds a pre-specified number, $M_I$, it indicates that the {BaRBF} is stuck in a local area.
Instead of continuing the search, we add some additional points to explore the experiment region. The additional point is chosen based on the maximin-distance criterion in the region.
That is, given the current explored points, we find the point such that the union set of this point with the current explored points has the maximal value of the minimal distance between any two points in this union set.
The purpose is to put additional points in the unexplored area as far away as possible from the existing points. We continue adding points until we obtain a better function value or until we add $M_T$ points. Then we will return to the original search procedure.
A similar idea has been adopted in \cite{regis2007stochastic} to detect if the search algorithm converges to a local optimum. We refer to this modificcation as {M-BaRBF}.}

\begin{table}[t] \caption{{Summary of optimal values obtained by M-{BaRBF} with Branin function. The run size of the initial design is 16, and the optimal value of the Branin function is 1.0473. \label{2-dim-M}}}
{\small
\begin{center}
\begin{tabular}{|c|l|l|l|l|l|l|l|l|}
  \hline
  Approach & 5\%   & Q1 & Median & Q3 & 95\%  & Mean & std & Frequencies with \\
           & Quantile&    &        &    & Quantile&      &     &true optimal values \\ \hline
  M-{BaRBF}(SEI) & 1.0397&1.0464 &1.0473&1.0473&	1.0473& 1.0458 & 0.0028 &38/60 \\ \hline
\end{tabular}
\end{center}}
\end{table}

{We implement the {M-BaRBF} for the Branin function by setting $M_I = M_T = 3$.
For the same initial points sets and other tuning parameters, the average performance of the 60 replicates for {M-BaRBF} is shown in Table \ref{2-dim-M}. Compare with the results for {BaRBF} in Table \ref{2-dim}. Except for the 5\% sample quantile value, the {M-BaRBF} performs better in the mean value and median value.
In addition, the frequency to reach the global optimum is 38/60 which is much higher than 29/60 for the {BaRBF}. The Q1 value for {M-BaRBF} is significantly higher than that for {BaRBF} and the median value touches the global maximum of the Branin function.
}

\subsection{{The Effects of the Grid Size}}

In the numerical examples, we suggest to pre-specify the grid size first when we implement our algorithm. This size may be chosen based on the prior knowledge. In the numerical examples in Section 4, the grid size is fixed as 0.04 or 0.05. Suppose we can consider different grid sizes for a given optimization problem. We will illustrate the effects of the grid sizes {on the performance.}

We revisit the 2D Branin function example in Section 4.1. Instead {of} setting {the} grid size as 0.04, 
we choose the finer size 0.02 and divide the region into $(51)^2$ grid points which still cover the original grid, $[0, 0.04, \dots, 1]^2$. {Based} on this finer grid, there are still two local maxima and the global optimal point is located at $[0.96, 0.16]$. {Since the number of grid points is now about four times that of the original, we take} more iterations, $4 \times 30 = 120$, for the proposed BaRBF. Then based on the same initial points and tuning parameters, the results with 60 replications are summarized in Table \ref{2-dim-f}. In this table, in addition to the results with 120 iterations, we also report the results with 30 iterations {for} comparison purpose.

\begin{table}[t] \caption{Summary of the 60 replications for the optimal values obtained by {BaRBF} with grid size, 0.02, in the 2-dimensional  experiment with Branin function. The run size of the initial design is 16, and the optimal value of the Branin function is 1.0473.} \label{2-dim-f}
{\small
\begin{center}
\begin{tabular}{|c|l|l|l|l|l|l|l|l|}
  \hline
  Approach & 5\%   & Q1 & Median & Q3 & 95\%  & Mean & std & Frequencies with \\
           & Quantile&    &        &    & Quantile&      &     &true optimal values \\ \hline
  {BaRBF}(120)& 1.0471 & 1.0471& 1.0473 &  1.0473 &  1.0473 &  1.0472 & 9.508e-5& 40/60\\ \hline
  {BaRBF}(30) & 1.0466 & 1.0471 &  1.0471 & 1.0473 & 1.0473 & 1.0469& 9.412e-4& 17/60 \\ \hline
\end{tabular}
\end{center}}
\end{table}

Compare the performances of {BaRBF} and {BaRBF}(120) in Tables \ref{2-dim} and \ref{2-dim-f} respectively. {First,} the {BaRBF}(120) is implemented over the finer grid  and it has higher frequency, $40/60$, to reach the global optimal point. In addition, the $5\%, 25\%$ quantiles and {the} median value of the best solutions of {BaRBF}(120) are 1.0471, 1.0471 and 1.0473 {respectively} which are significantly higher than { the corresponding values shown in Table \ref{2-dim}.} 
Obviously {BaRBF}(120) has the higher mean value 1.0472 and {a} smaller standard deviation. This may be related to {the fact} that there are more candidate points and larger number of iterations. Thus {BaRBF} can identify better function values due to finer grid and {can still} explore the experimental space because of a larger number of iterations. To {support this guess}, we also report the summary of the {BaRBF} with finer grid and 30 iterations, denoted by {BaRBF}(30). The corresponding $5\%$, $25\%$ sample quantiles and {the} median value are still better than {the corresponding values shown in Table \ref{2-dim}.} But the frequency for obtaining the global optimal point is only 17/60. It means that 30 iterations may not be large enough for {BaRBF} to explore the whole region and the search process may get stuck in some local areas. Thus we need to have more iterations to increase the probability to jump out {of} these local areas. Overall we can conclude that {when} we have {a} finer grid, {a larger} number of iterations should be necessary.

\subsection{{A  Grid-free Method}}

When we demonstrate the performance of the proposed {BaRBF} in Section 4, we mention that we need to pre-specify a grid as the explored candidate point set before implementing the {BaRBF}. For example, we set the grid size as 0.04 for {the} 2 and 3 dimension cases. {Suppose we consider} the application in parameter tuning. In practice, the precision of each variable should be limited and due to this grid set-up, we can easily select the next explored point based on the SEI criterion; otherwise it should be treated as another non-differentialable optimization problem.
However, the problem for the grid set is the curse of  dimensionality. For example, when we consider a 7-dimension region, $[0, 1]^7$. Suppose we choose the grid size as 0.2. {Then} the number of grid points is $6^7 = 279936$ points, which is huge. 
To keep such huge number of grid points in our process {would slow down} the computational speed. {Since the grid point set is used as the} candidate set for selecting the next explored point, we would propose another modified {BaRBF} without pre-specifying grid point set.

The idea is similar to the {G-MSRBF} in \cite{regis2007stochastic}. {Instead of} fixing the grid points for the candidate point set, we randomly generate the candidate points in each iterations. \cite{regis2007stochastic} have mentioned two conditions for this point generation procedure. {One is the conditionally independent sampling condition and another one is the dense condition, that is, any point in $\chi$ can be chosen as a candidate point with a positive probability. Since our approach satisfies the first condition,} 
{sample} the candidate points from the uniform distribution over the experimental region $\chi$ {to satisfy the dense condition}
{to} guarantee the convergence property, i.e., when $N$ goes to infinity, the global optimal point can be identified by such search process with high probability. Thus in our algorithm, we add one {more} step to generate the candidate point set $\chi_N$ from the uniform distribution over the region and then {choose the} next explored points by {applying the} SEI criterion to $\chi_N \setminus P_{exp}$. After updating the $P_{exp}$, we would continue the selection process by re-generating the new candidate points.

\begin{figure}
\centering
\includegraphics[width=5.5in]{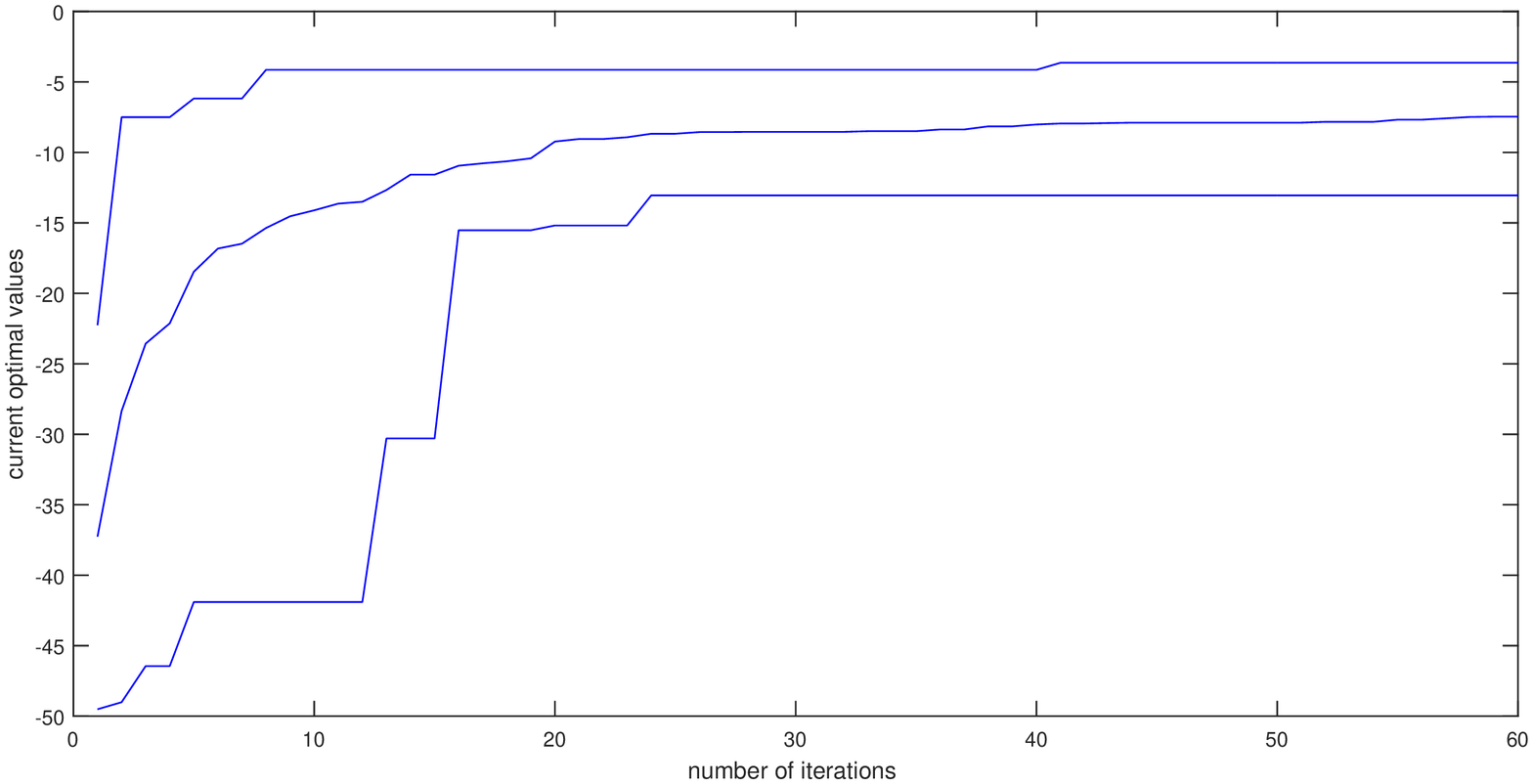}
 \protect\caption{
 Three lines are the 5\% quantile; mean value and 95\% quantile of current optimal values based on 30 replications for the example of Rastrigin function.} \label{fig:rastrigin}
\end{figure}

To demonstrate this new modified {BaRBF}, named the grid-free {BaRBF}, the Rastrigin function \citep{Poh:2015} is taken as the objective function, i.e.,
$$f(\mathbf{x}) = - 10 d - \sum_{i =1}^{d} [(x_i-0.5) - 10 \cos(2\pi (x_i-0.5))].$$
This Rastrigin function has multiple local {maxima} and the locations of the local maximal points are uniformly distributed. The global optimal point is $[0.5, \dots, 0.5]$ with maximum value 0. Here we choose $d = 8$, that is, we consider {a} 8-dimensional problem. At first the 80 initial points are chosen from a Latin hypercube design over the experimental region, $\chi = [0,1]^8$. The number of iteration for grid-free {BaRBF} is 60 and at each iteration, we uniformly sample 8,000 points from $\chi$. The results based on 30 independently replications are summarized in Table \ref{8-dim}. Figure \ref{fig:rastrigin} shows the $5\%$, $95\%$ quantile curves and the mean values with respect to the number of iterations. Overall the results {suggest} that the proposed grid-free {BaRBF} does improve the objective
function values over the whole search process; especially {in} the first few iterations, the {improvement is}  significant. 
{However, the improvement does slow down later.}
In addition 
we also {tested} the case of the Rastrigin function with $d = 10$ and the results share {a similar} pattern. Thus we do not show the corresponding results here to save space. The G-MSRBF is also implemented for this Rastrigin function example. {Following the same set-up for generating the candidate points, the results with 30 independent replications are also summarized in Table \ref{8-dim}. The proposed Grid-free BaRBF performs better than the G-MSRBF, especially in terms of the three sample quantiles and the mean values.}  

\begin{table}[t]
\caption{Summary of maximum values obtained by grid-free {BaRBF}
with 30 replications in the 8-dimensional case. The optimal value of
the 8-dimensional Rastrigin function is 0.0000.} \label{8-dim}
\begin{center}
\begin{tabular}{|c|l|l|l|l|l|l|}
  \hline
  Approach & Q1 & Median & Q3 & Mean & std & Maximal value \\ \hline
  {Grid-free {BaRBF}} & -9.0985 & -6.5520 & -5.2748 &-7.4002 & 2.9071 & -3.5842 \\ \hline
  G-MSRBF        &-9.4032 & -7.3887 & -5.9366 & -7.6108 & 2.4987 & -3.1462 \\ \hline
\end{tabular}
\end{center}
\end{table}

\section{Conclusion and Future Work} \label{sec:conclusion}

We have proposed a global optimization framework that utilizes an
adaptive RBF-based Bayesian surrogate model to approximate the true
function, and to guide the selection of  new points for function
evaluation. There is novelty in both steps of the strategy. First,
we use a hierarchical normal mixture surrogate model, where the
parameters in the RBFs can be automatically updated to best
approximate the true function. Second, 
the sample EI criterion is employed as a selection criterion.
We have conducted some extensive numerical studies 
on standard test functions.
The results demonstrate that the proposed {BaRBF} is more efficient and stable for searching the global maximizer compared with the {G-MSRBF}.
{For the comparison between the EGO and the {BaRBF}, their performance depends on the characterization of the true objective functions.
Since the Branin function and the 4D Hartmann function are quit smooth, the EGO has better performances.
For the 2D and 3D Ronkkonen functions, the functions are not smooth and have more multiple extreme points.
The {BaRBF} outperforms the EGO.}

There are some directions for future research. First, the point
selection criterion is a key element of {BaRBF}.
A good selection criterion is
to balance the trade-off between {exploitation and exploration.}
Currently the Sampled EI criterion is adopted in the {BaRBF}. When the Gaussian prediction assumption is held, the EI criterion is a weighted sum of the prediction improvement and prediction variation which can be treated as a way to balance the exploitation and exploration.
However, in the {BaRBF}, we do not have the distribution assumption and thus how SEI to balance these two properties is uncertain.
Based on the numerical results in two 2D functions, our SEI criterion tends to have the exploitation property but less effect related to the exploration,
because for the smooth Branin function, {BaRBF} may be stuck in a local area, but {BaRBF} can quickly identify the global maximum close to the explored points in the example of Ronkkonen function.
Thus one possibility is to add the prediction variation measurement, i.e., to
quantify the prediction uncertainties by the 95\% confidence interval bandwidth of $f_N(\bf x)$:
\begin{eqnarray}   \label{eq: CIB}
CIB(f_N({\bf x})) = UCI(f_N({\bf x})) - LCI(f_N({\bf x})),
\end{eqnarray}
where $UCI(f_N({\bf x})), LCI(f_N({\bf x}))$ are the upper CI and lower CI calculated as the 97.5\% and 2.5\% quantiles of the posterior samples $f_N^{(k)}({\bf x}).$ However, the problem should be how to integrate the SEI and CIB together.
In
addition, there should be a data-driven tuning procedure for
hyper-parameters in {BaRBF}, e.g., the value of $C$ in the mixture
normal prior. 
The tuning procedure suggested in
\cite{chen2016sequential} {might serve as a starting point.}

In {BaRBF}, {to simplify our algorithm, we fix the center parameters $\mu_i$ as the explored points $\mathbf{x}_i$. Thus it only depends on the current $N$ explored points and is independent of the dimension of the region $V$ and the resolution of the grid.}
Since we need to pre-specify the candidate point set for next point selection, in this paper, we fix the candidate points as the
whole grid. However, when the number of grid points is {large}, it
will not be efficient to consider the whole grid. 
{Instead we propose a grid-free version by iteratively generating the candidate points from uniform distribution over the experimental region,}
and then implement the procedure over this
candidate set to identify next explored points. {Based on the theoretical results in \cite{regis2007stochastic}, the convergence property {can hold} when the number of explored points goes to infinity.}
{In addition,}
to identify the true optimal point, an
adaptive {BaRBF} method can {be considered as follows.} At each
iteration, we refine the current grid locally based on the hot spot
areas identified from the surrogate surface, and then re-run {BaRBF} in
these local areas independently. Take the Branin function example in
Figure \ref{fig:1_brainin} {for} illustration. In Figure
\ref{fig:1_brainin}(f), we can identify three hot spot areas. Then
we can choose three smaller disjoint regions with finer grid to
cover each hot spot area, and then individually implement {BaRBF} for
each region. We can continue this procedure until the grid size in
each region is small enough.
{In fact, this adaptive procedure is similar to the Multistart Local
MSRBF (ML-MSRBF) in \cite{regis2007stochastic} 
{and the two versions can be integrated.} 
Following {the} ML-MSRBF, we can treat the area around the current best value as the local hot spot and 
restart the BaRBF by choosing candidates from a normal distribution centered at the best point in this local hot spot with a certain variance. After {some number of} iterations, we may identify another best point and restart {the} BaRBF again. Repeat this procedure until the stopping criterion is met.}

{Finally we mention a possible approach for tuning the value of $C$ in our proposed procedure. In Section 3, we give the guideline that we should choose $C$ larger than 0. But how large is the value of $C$? According to the Bayesian variable selection literature, $C$ can be tuned via the cross validation approach. {Based  on this approach, we can consider} the following suggestion. After collecting the initial observations, we may tune the value of $C$ via cross validation and keep this value for a certain number of iterations. Then we can re-tune the value of $C$ via cross validation again.}

\begin{acknowledgements}
{ Chen's research is supported by Ministry of Science and Technology (MOST) of Taiwan 104-2918-I-006-005 and the Mathematics Division of
the National Center for Theoretical Sciences in Taiwan. Wu's research is supported by ARO W911NF-17-1-0007 and NSF DMS-1564438.}
\end{acknowledgements}

\bibliographystyle{apalike}
\bibliography{referencefile}

\begin{thebibliography}{}

\bibitem[Andrieu et~al., 2001]{andrieu2001robust}
Andrieu, C., De~Freitas, N., and Doucet, A. (2001).
\newblock Robust full bayesian learning for radial basis networks.
\newblock {\em Neural Computation}, 13(10):2359--2407.

\bibitem[Bishop, 2006]{bishop2006pattern}
Bishop, C.~M. (2006).
\newblock {\em Pattern Recognition and Machine Learning}.
\newblock New York: NY, Springer.

\bibitem[Boyd and Vandenberghe, 2004]{boyd2004convex}
Boyd, S. and Vandenberghe, L. (2004).
\newblock {\em Convex Optimization}.
\newblock New York: NY, Cambridge University Press.

\bibitem[Buhmann, 2003]{buhmann2003radial}
Buhmann, M.~D. (2003).
\newblock {\em Radial Basis Functions: Theory and Implementations}, volume~12.
\newblock New York: NY, Cambridge University Press.

\bibitem[Chen et~al., 2011]{chen2010building}
Chen, R.~B., Wang, W., and Wu, C. F.~J. (2011).
\newblock Building surrogates with overcomplete bases in computer experiments
  with applications to bistable laser diodes.
\newblock {\em IIE Transactions}, 43(1):39--53.

\bibitem[Chen et~al., 2017]{chen2016sequential}
Chen, R.-B., Wang, W., and Wu, C. F.~J. (2017).
\newblock Sequential designs based on bayesian uncertainty quantification in
  sparse representation surrogate modeling.
\newblock {\em Technometrics}, 59(2):139--152.

\bibitem[Chipman et~al., 1997]{chipman1997bayesian}
Chipman, H., Hamada, M., and Wu, C. F.~J. (1997).
\newblock A bayesian variable-selection approach for analyzing designed
  experiments with complex aliasing.
\newblock {\em Technometrics}, 39(4):372--381.

\bibitem[Chipman et~al., 2012]{chipman2012}
Chipman, H., Ranjan, P., and Wang, W. (2012).
\newblock Sequential design for computer experiments with a flexible bayesian
  additive model.
\newblock {\em The Canadian Journal of Statistics}, 40:663--678.

\bibitem[Fasshauer and Zhang, 2007]{fasshauer2007choosing}
Fasshauer, G.~E. and Zhang, J.~G. (2007).
\newblock On choosing ``optimal'' shape parameters for rbf approximation.
\newblock {\em Numerical Algorithms}, 45(1-4):345--368.

\bibitem[George and McCulloch, 1993]{george1993variable}
George, E.~I. and McCulloch, R.~E. (1993).
\newblock Variable selection via gibbs sampling.
\newblock {\em Journal of the American Statistical Association},
  88(423):881--889.

\bibitem[Gutmann, 2001]{gutmann2001radial}
Gutmann, H.~M. (2001).
\newblock A radial basis function method for global optimization.
\newblock {\em Journal of Global Opt}, 19(3):201--227.

\bibitem[Jones, 2001]{jones2001taxonomy}
Jones, D.~R. (2001).
\newblock A taxonomy of global optimization methods based on response surfaces.
\newblock {\em Journal of Global Opt}, 21(4):345--383.

\bibitem[Jones et~al., 1998]{jones1998efficient}
Jones, D.~R., Schonlau, M., and Welch, W.~J. (1998).
\newblock Efficient global optimization of expensive black-box functions.
\newblock {\em Journal of Global Opt}, 13(4):455--492.

\bibitem[Koutsourelakis, 2009]{koutsourelakis2009accurate}
Koutsourelakis, P.~S. (2009).
\newblock Accurate uncertainty quantification using inaccurate computational
  models.
\newblock {\em SIAM Journal on Scientific Computing}, 31(5):3274--3300.

\bibitem[Mockus et~al., 1978]{Mockus1978}
Mockus, J., Tiesis, V., and Zilinskas, A. (1978).
\newblock The application of bayesian methods for seeking the extremum.
\newblock {\em Towards Global Optimization}, 2:117--129.

\bibitem[Morris and Mitchell, 1995]{morris95}
Morris, M.~D. and Mitchell, T.~J. (1995).
\newblock Exploratory designs for computer experiment.
\newblock {\em J. Statist. Plann. Inference}, 43:381--402.

\bibitem[Oefelein and Yang, 1998]{oefelein1998}
Oefelein, J.~C. and Yang, V. (1998).
\newblock Modeling high-pressure mixing and combustion processes in liquid
  rocket engines.
\newblock {\em Journal of Propulsion and Power}, 14(5):843--857.

\bibitem[Picheny et~al., 2013]{picheny2013}
Picheny, V., Wagner, T., and Ginsbourger, D. (2013).
\newblock A benchmark of kriging-based infill criteria for noisy optimization.
\newblock {\em Structural and Multidisciplinary Optimization}, 48(3):607--626.

\bibitem[Pohlheim, 2015]{Poh:2015}
Pohlheim, H. (2015).
\newblock Geatbx examples: Examples of objective functions.
\newblock \url{http://www.geatbx.com/download/GEATbx_ObjFunExpl_v37.pdf}.

\bibitem[Regis and Shoemaker, 2007]{regis2007stochastic}
Regis, R.~G. and Shoemaker, C.~A. (2007).
\newblock A stochastic radial basis function method for the global optimization
  of expensive functions.
\newblock {\em INFORMS Journal on Computing}, 19(4):497--509.

\bibitem[R\"{o}nkk\"{o}nen et~al., 2011]{ronkkonen2011}
R\"{o}nkk\"{o}nen, J., Li, X., and Kyrki, V. (2011).
\newblock A framework for generating tunable test function for multimodal
  optimization.
\newblock {\em Soft Comput}, 15:1689--1706.

\bibitem[Santner et~al., 2013]{santner2013design}
Santner, T.~J., Williams, B.~J., and Notz, W.~I. (2013).
\newblock {\em The Design and Analysis of Computer Experiments}.
\newblock Springer Science.

\bibitem[Torn and \v{Z}ilinskas, 1989]{Torn:1989:GO:75021}
Torn, A. and \v{Z}ilinskas, A. (1989).
\newblock {\em Global Optimization}.
\newblock Springer-Verlag, Berlin, Heidelberg.

\bibitem[\v{Z}ilinskas, 2010]{zilinskas2010}
\v{Z}ilinskas, A. (2010).
\newblock On similarities between two models of global optimization:
  statistical models and radial basis functions.
\newblock {\em Journal of Global Opt}, 48:173--182.

\bibitem[\v{Z}ilinskas and Zhigljavsky, 2016]{zilinskas2016}
\v{Z}ilinskas, A. and Zhigljavsky, A. (2016).
\newblock Stochastic global optimization: a review on the occasion of 25 years
  of informatica.
\newblock {\em Informatica}, 27:229--256.

\bibitem[Zellner, 1986]{zellner1986assessing}
Zellner, A. (1986).
\newblock On assessing prior distributions and bayesian regression analysis
  with g-prior distributions.
\newblock {\em Bayesian Inference and Decision Techniques: Essays in Honor of
  Bruno De Finetti}, 6:233--243.

\end{thebibliography}

\end{document}